\documentclass[10pt,journal,compsoc,UTF8]{IEEEtran}

\ifCLASSOPTIONcompsoc
  \usepackage[nocompress]{cite}
\else
  \usepackage{cite}
\fi
\ifCLASSINFOpdf
\else
\fi
\usepackage{epsfig,xspace,url,xspace,color,multirow}
\usepackage{graphicx,amssymb,amsmath,amsthm}
\usepackage{epstopdf}
\usepackage{times}
\usepackage{subfigure}
\usepackage{caption}
\usepackage{flushend}
\usepackage{booktabs}
\usepackage[justification=centering]{caption}
\usepackage{latexsym}
\usepackage{amsmath}
\usepackage{amssymb}
\usepackage{arydshln}
\usepackage{stfloats}
\usepackage{cite}
\usepackage[square, comma, sort&compress, numbers]{natbib}
\usepackage[switch]{lineno}
\usepackage[boxed, linesnumbered, ruled]{algorithm2e}

\def\BibTeX{{\rm B\kern-.05em{\sc i\kern-.025em b}\kern-.08em
    T\kern-.1667em\lower.7ex\hbox{E}\kern-.125emX}}

\usepackage{multirow}
\usepackage[normalem]{ulem}
\useunder{\uline}{\ul}{}
\usepackage[T1]{fontenc}

\usepackage{float}
\usepackage{threeparttable}
\usepackage[table,xcdraw]{xcolor}

\usepackage{color}
\usepackage {ulem}

\makeatletter
\newcommand{\removelatexerror}{\let\@latex@error\@gobble}
\makeatother

\makeatletter

\newcommand{\Rmnum}[1]{\expandafter\@slowromancap\romannumeral #1@}
\makeatother

\newcommand{\oldstuff}[1]{}
\definecolor{gray}{rgb}{0.5,0.5,0.5}

\renewcommand\thesection{\arabic{section}}

\begin{document}
\title{GNN-Geo: A Graph Neural Network-based Fine-grained IP geolocation Framework}

\author{Shichang~Ding,
        Xiangyang~Luo*,
        Jinwei~Wang,~\IEEEmembership{Member,~IEEE,}% <-this % stops a space
        ~Xiaoming~Fu,~\IEEEmembership{Fellow,~IEEE}
\thanks{\textbf{This article has been accepted for publication in IEEE Transactions on Network Science and Engineering. This is the author's version which has not been fully edited and content may change prior to final publication. Citation information: DOI 10.1109/TNSE.2023.3266752}}
%\thanks{This work has been submitted to the IEEE for possible publication. Copyright may be transferred without notice, after which this version may no longer be accessible.}
\IEEEcompsocitemizethanks{\IEEEcompsocthanksitem Shichang~Ding and Xiangyang~Luo are with State Key Laboratory of Mathematical Engineering and Advanced Computing, Zhengzhou 276800, China.\emph{(Corresponding author: Xiangyang Luo)}\protect\\
E-mail: scdingwork@outlook.com, luoxy\_ieu@sina.com
\IEEEcompsocthanksitem Jinwei~Wang is with School of Computer and Software, Nanjing University of Information Science \& Technology, Nanjing 210044, China.\protect\\
E-mail: wjwei\_2004@163.com% <-this % stops an unwanted space
\IEEEcompsocthanksitem Xiaoming~Fu is with Institute of Computer Science, University of G\"ottingen, G\"ottingen 37077, Germany.\protect\\
E-mail: fu@cs.uni-goettingen.de}% <-this % stops an unwanted space
\thanks{Manuscript received ~, ~; revised ~, ~.}}

\markboth{IEEE Transactions on Network Science and Engineering, ~Vol.~, No.~, ~}%
{Ding \MakeLowercase{\textit{et al.}}: GNN-Geo: A Graph Neural Network-based Fine-grained IP geolocation Framework}

\IEEEtitleabstractindextext{%
\begin{abstract}
Rule-based fine-grained IP geolocation methods are hard to generalize in computer networks which do not follow hypothetical rules. Recently, deep learning methods, like multi-layer perceptron (MLP), are tried to increase generalization capabilities. However, MLP is not so suitable for graph-structured data like networks. MLP treats IP addresses as isolated instances and ignores the connection information, which limits geolocation accuracy. In this work, we research how to increase the generalization capability with an emerging graph deep learning method -- Graph Neural Network (GNN). First, IP geolocation is re-formulated as an attributed graph node regression problem. Then, we propose a GNN-based IP geolocation framework named GNN-Geo. GNN-Geo consists of a preprocessor, an encoder, messaging passing (MP) layers and a decoder. The preprocessor and encoder transform measurement data into the initial node embeddings. MP layers refine the initial node embeddings by modeling the connection information. The decoder maps the refined embeddings to nodes' locations and relieves the convergence problem by considering prior knowledge. The experiments in \textcolor{black}{8 real-world IPv4/IPv6 networks in North America, Europe and Asia show the proposed GNN-Geo clearly outperforms the state-of-art rule-based and learning-based baselines}. This work verifies the great potential of GNN for fine-grained IP geolocation.
\end{abstract}

}
\maketitle
\IEEEdisplaynontitleabstractindextext
\IEEEpeerreviewmaketitle

\IEEEraisesectionheading{\section{Introduction}}

    \label{sec:intro}
    \IEEEPARstart{I}{P} geolocation is to obtain the geographic location (geolocation) of an IP address \cite{li2021geocam,ding2021street}. It is widely used in localized online advertisement, location-based content restriction, cyber-crime tracking, and so on \cite{li2021geocam,wang2020towards,DanPD22}. IP geolocation is important especially for networking devices without GPS (Global Positioning System) function, such as PCs and routers. IP geolocation can be categorized into two kinds by accuracy: coarse-grained IP geolocation and fine-grained IP geolocation. Coarse-grained IP geolocation can find the city or state location of an IP address. Researchers have proposed a series of coarse-grained IP methods such as \cite{wang2020towards,li2021geocam}. The median error distances of these methods are usually between dozens and several hundreds of kilometers. After obtaining a coarse-grained location of a target IP, fine-grained IP geolocation methods can be used to find its more accurate location. The median error distances of fine-grained IP geolocation methods are usually less than 10 kilometers. Though there are several fine-grained IP geolocation methods like SLG \cite{SLG}, Checkin-geo \cite{liu2014mining}, Corr-SLG \cite{ding2021street} and MLP-Geo \cite{zhang2020geolocation}, they stills face many challenging problems. \textbf{In this paper, we mainly focus on improving the generalization capabilities of measurement-based fine-grained IP geolocation in different computer networks}.\\
    \indent
    Measurement-based IP geolocation can geolocate a target IP after the probing host \cite{ding2021street} obtains the measurement data (e.g., delay) to the target IP and landmarks \cite{SLG,ding2021street}. Landmarks are IP addresses with known locations \cite{ding2021street}. Generally, existing measurement-based fine-grained IP geolocation methods can be categorized into two kinds: rule-based and deep learning-based methods (referred as learning-based methods in this paper). For rule-based methods like SLG \cite{SLG} and Corr-SLG \cite{ding2021street}, they assume that most hosts in one computer network are following some delay-distance rules. These hypothetical rules are proposed by human experts based on their observations. Then a target IP can be mapped to its geolocation based on the delay-distance rules. The performances of rule-based methods rely on the partition of hosts which follow the hypothetical rules. \textbf{This limits the generalization capabilities of rule-based methods in networks where hosts don't follow their hypothetical rules}.\\
    \indent
    Recently, deep learning seems to be a hopeful solution to the generalization problem of IP geolocation methods. There are two possible advantages of learning-based IP geolocation. First, learning-based methods do not rely on hypothetical rules which are suitable for all computer networks. They can learn a "measurement data - locations" relationship suitable to a specific computer network based on this network's raw measurement data. Second, learning-based methods are capable of extracting non-linear relationships, while non-linear rules are hard for human experts to observe. Encouraged by these possible advantages, researchers proposed several IP geolocation methods based on MLP (multi-layer perceptron), such as NN-Geo \cite{jiang2016ip} and MLP-Geo \cite{zhang2020geolocation}. Compared with NN-Geo, MLP-Geo is more accurate because it leverages not only delay but also routing information.\\
    \indent
    However, it is hard for MLP to model the connection information of computer networks. MLP is not good at modeling non-Euclidean structured data. Nevertheless, computer networks are fundamentally represented as graphs, a typical non-Euclidean structured data \cite{9046288,dwivedi2020benchmarking}. MLP can only treat IP addresses as isolated instances, though hosts are non-independent and linked to each other through intermediate routers. This is why MLP-Geo can only leverage the router IDs (and the delay) between the probing host and the targets, ignoring all the other useful features, such as the links and delays between routers. Based on the experiments (shown in Table \ref{tab:mainresults}), MLP-Geo could be less accurate than rule-based methods in some actual networks. \textbf{This indicates that IP geolocation needs a deep learning method more suitable for graph-structured data}.\\
    \indent
    Graph Neural Network (GNN) is an emerging deep learning method for graph-structured data presentation \cite{1st_GCN}. Recently, it has been successfully applied in different fields of graph-structured data, like recommendation system \cite{wu2021self}, social network \cite{2019Graph}, knowledge map \cite{wu2021disenkgat}, etc. However, little attention has been paid to introducing GNN into IP geolocation. As far as we know, this is the first work to explore the potential of GNN in fine-grained IP geolocation. The main challenges of introducing GNN into IP geolocation tasks are \textcolor{black}{at least} three-folds.
    \begin{enumerate}
    \item How to formulate IP geolocation problem in the context of graph deep learning?
    \item How to design a GNN-based Framework for the fine-grained measurement-based IP geolocation task?
    \item \textcolor{black}{Which factors would clearly affect the performance of GNN in fine-grained IP geolocation performance?}
    \end{enumerate}

    \indent
    By tackling these challenges, the main contributions of our work can be summarized as follows.
    \begin{enumerate}

    \item \textbf{Formulating IP geolocation as a graph learning task}. We formulate the measurement-based fine-grained IP geolocation task as a semi-supervised attributed-graph node regression problem. First, IP addresses are mapped into graph nodes, and links are mapped into graph edges. Then, the features of IP addresses (e.g., delay) and links (e.g., the delay of link) are transformed into attributes of the graph nodes and edges. Finally, the IP geolocation problem is formulated as estimating the latitudes and longitudes of the target nodes based on the known locations of other nodes as well as the topology and attributes of the graph. This paves the road of introducing graph deep learning methods into IP geolocation.

    %\item \textcolor{black}{sout{\textbf{Theoretical analysis for GNN's advantages in IP geolocation}. By analyzing the difference in forming the embeddings of IP addresses, we theoretically explain the advantages of proposed GNN-based method compared to previous learning-based (MLP) method in IP geolocation.}}

    \item \textbf{GNN-Geo framework}. We propose a GNN-based framework (GNN-Geo) to solve the node regression problem. It consists of four components: a preprocessor, an encoder, MP (message passing) layers, and a decoder. The preprocessor transforms the raw traceroute data of a computer network into an attributed-graph representation. The encoder generates the initial graph node/edge embeddings for MP layers. MP layers refine the node embeddings by propagating messages along the edges. Finally, the latitudes and longitudes of each node are decoded from the refined node embeddings.

	\item \textbf{\textcolor{black}{Improving MLP-based decoder with rules}}. We highlight the limitation of the vanilla \textcolor{black}{MLP-based} decoder, which makes GNN model hard to converge. The convergence problem is relieved with \textcolor{black}{an idea like "rule-based" methods:} making GNN-Geo to find targets' fine-grained locations inside a possible coarse-grained area \textcolor{black}{instead of the whole earth}. Experiments show that this significantly reduces the error distances and training epochs of GNN-Geo. Take Hong Kong as an example, compared with the vanilla \textcolor{black}{MLP-based} decoder, the \textcolor{black}{improved} decoder reduces the error distance from more than 2,690 km to less than 8 km.
	 	
    \item \textcolor{black}{\textbf{Large-scale experiment results outperform the best baselines}. The experiments are carried out in 8 real-world IPv4/IPv6 networks including New York State, Hong Kong, Shanghai, Beijing, Tokyo, Tokyo (IPv6), Berlin (IPv6) and Munich (IPv6). The results show: (i) when the training dataset ratio is 70\%, GNN-Geo outperforms the best baselines averagely by more than 21\% w.r.t. both average error distance and median error distance; (ii) When training dataset ratio decrease from 70\% to 10\%, though its advantages are reduced on sparser dataset, GNN still outperform baselines in most cases. We also analyze how different factors (e.g., basic GNN models, GNN aggregation methods) would affect GNN-Geo's performance. These results validate the value of introducing GNN in fine-grained IP geolocation.}
    \end{enumerate}

    The rest of the paper is organized as follows. Related works are introduced in Section \ref{rw}. In Section 3, IP geolocation is formulated into a graph node regression problem. \textcolor{black}{Then in Section 4, we discuss why GNN is worth to be tried in IP geolocation.} In Section \ref{Model}, we introduce the detail of proposed GNN-Geo. Section \ref{sec:evaulation} shows and analyzes the experiment results in three real-world datasets. Section \ref{sec:discussion} discusses several possible future improvements for GNN-Geo. Finally, the paper is concluded in Section \ref{sec:conclusion}.

\section{related work}
    \label{rw}
    In this section, we introduce the works related to our study, mainly including fine-grained IP geolocation and Graph Neural Network.

    \subsection{Fine-grained IP geolocation}
    \label{rw_ipgeo}

    Besides measurement-based methods, there are two other kinds of fine-grained IP geolocation methods: database and data-mining-based method. IP geolocation databases can only provide the fine-grained locations for a small proportion of IP addresses, which could hardly satisfy people's needs \cite{dan2021ip}. Data-mining-based methods, like Checkin-Geo \cite{liu2014mining}, can map IP addresses to locations based on a large amount of raw data which contains "IP-location" information such as users' logs. However, these kinds of raw data resources are only possessed by few internet giants like Microsoft Bing \cite{dan2021ip,DanPD22} and Tencent \cite{liu2014mining}, which are not open to the public. For example, Checkin-Geo is a data-mining-based fine-grained method. First, it extracts users' names and their locations from the checkin \cite{liu2014mining} information collected on smartphone applications. Then it extracts users' names and users' IP addresses from the logins collected on PC applications. In the end, Checkin-Geo can map a user's PC IP address to his/her smartphone location. Both the PC and the smartphone applications are owned by a famous Chinese online social platform called "Tencent QQ". Recently, researchers from Microsoft build a large-scale IP-location ground-truth dataset, which is mined from the query logs of a commercial search engine. Based on the large-scale dataset, they can use machine learning methods to estimate the location of IP addresses from their reverse DNS hostnames \cite{DanPD22}. We can see, these high quality large-scale ground truth data resources are hard to get for researchers outside Tencent or Microsoft. Besides this problem, these methods cannot work well in areas where those Internet giants are not serving, or the location-sharing service is forbidden. For example, Tencent QQ mainly serves in China and has been banned in India\footnote{www.xda-developers.com/india-bans-xiaomi-mi-browser-pro-qq-international-app/}.
    \\
    \indent
    The limitations of database and data-mining-based methods require researchers to pay attention to measurement-based methods. In theory, measurement-based methods can geolocate any target IP if the delay or routing data between the target IP address and landmarks can be obtained. However, it is hard for measurement-based methods to get a fine-grained geolocation result. Because the relationship between measurement data and target IP's location is very complicated and different in various computer network environments.\\%Though there have been several measurement-based fine-grained methods, their performances vary significantly in different network environments. How to design a fine-grained method that can well generalize in different networks is still an important problem.\\
    \indent
    SLG is the first fine-grained measurement-based IP geolocation method \cite{SLG}. It is also the first rule-based method as well as one of the most widely-used fine-grained IP geolocation baselines until now. It assumes that most hosts in a computer network are following a simple linear delay-distance rule: the shortest "relative delay" comes from the nearest landmark. Since the delay between a landmark and a target IP is hard to measure, "relative delay" is proposed by SLG as an approximation. Assume the delay from the probing host to a target IP is $d_{pt}$, the delay from the probing host to a landmark is $d_{pl}$, and the delay from the probing host to the closest common router \cite{SLG,ding2021street} shared by the target IP and landmark is $d_{pr}$, the "relative delay" between the target and the landmark is $(d_{pt}-d_{pr})+(d_{pl}-d_{pr})$. This rule originated from an observation of the relationship between relative delay and distance of 13 landmarks in New York City \cite{SLG}. However, in further studies, researchers realized that this rule may be not valid for all intra-city networks \cite{ding2021street}.\\
    \indent
    Some researchers find that there is no clear relationship between relative delay and distance in cities like Zhengzhou (China) and Toronto (Canada) \cite{ding2021street}. They explain that for the intra-city delay, the delay caused by geographical distance are not always the major constituent. The other factors like queuing delay and processing delay in routers could be more important. Thus the relationship between relative delay and distance is very complicated. In Corr-SLG, the relative-delay-distance correlation of all landmarks is calculated at first. Then all landmarks are divided into three groups. Corr-SLG leverages different delay-distance rules in each group. In the groups with a strong-positive relative-delay-distance correlation (close to 1), the relative delay increases as the distance increases. Thus the rule is similar to SLG, and Corr-SLG still map targets to the landmarks with the shortest relative delay. In the groups with a strong-negative correlation, they map targets to the landmarks with the largest relative delay (close to -1). However, Corr-SLG cannot decide how to handle the groups with weak correlation (close to 0). They simply map these targets to the average locations of landmarks. We can see the accuracy of the rule-based methods relies on the partition of hosts which follows their pre-assumed rules. Since the network environments may change in each city and at different times, \textcolor{black}{the relationships between delay and distance may vary significantly. Thus,} the performances of rule-based methods vary significantly. \textbf{How to design a fine-grained method that can well generalize in different networks is an important problem}.\\
    \indent
    In recent years, deep learning has shown clear advantages in various networking problems \cite{DBLP:journals/tnse/KumarKSGTGX21,DBLP:journals/tnse/GuoLODDX22,DBLP:journals/tnse/ChenZQXX22,DBLP:journals/tnse/XiaWTXW21,DBLP:journals/tnse/MaoZX21,nagaraj2022passenger}. People also begin to try deep learning methods in IP geolocation to improve the generalization capabilities. Both NN-Geo \cite{jiang2016ip} and MLP-Geo \cite{zhang2020geolocation} use MLP to estimate locations for target IP addresses. Compared with NN-Geo, MLP-Geo adds a new kind of useful information -- the routers' IDs between probing hosts and the targets. Its performance is clearly better than NN-Geo since NN-Geo only utilizes the delay between probing hosts and the target IP. The aim of learning-based methods is not to find a rule which is suitable for all networks. Instead, they aim to present a method which can correctly find "rules" in different networks. Whether a method can find rules correctly mainly relies on its modeling ability towards the target networks. Actually, MLP is not so suitable for modeling the measurement data of computer networks. MLP is good at prepossessing Euclidean structure data. However, the topology of the internet is a non-Euclidean structure data \cite{WeiweiJiang2021GraphbasedDL}. MLP can only treat target IP addresses as isolated data instances while ignoring the connection information between targets. This would easily lead to suboptimal representations and performance impairment. We still need to improve the learning-based methods by utilizing deep learning methods more suitable to networks.

    \subsection{Graph Neural Network (GNN)}
    \label{rw_gcn}
    Graph Neural Network (GNN) is an emerging deep learning method, which is specially designed for graph-structured data \cite{1st_GCN,velivckovic2017graph,gilmer2017neural}. GNN learns the representation of a graph node by recursively aggregating the information of its neighbor nodes. Then, the learned node representations can be used for node-level, edge-level or graph-level classification/regression tasks. There are several typical GNN architectures such as GCN (Convolutional Neural Network) \cite{1st_GCN}, GAT (Graph Attention Network) \cite{velivckovic2017graph}, MPNN (Message Passing Neural Network) \cite{gilmer2017neural}, etc. Most existing GNN-based methods can be seen as their variants or improved version. GCN is inspired by CNN's success in image processing area \cite{long2015fully}. GCN uses multiple graph convolutional layers to aggregate information from neighbor nodes. GAT introduces attention mechanisms into information aggregation process. GCN usually aggregates the normalized sum of neighbors' information while GAT assigns attention weights to neighbors to get a weighted sum of neighbors' information. This can help GAT to differentiate the contributions of different neighbors and gain better performance. Both GCN and GAT mainly focus on aggregating node feature and ignore edge feature \cite{chen2019utilizing}. However, edge feature is crucial for IP geolocation tasks. For example, the delay or hop count between IP addresses have been verified as very important features in IP geolocation in previous methods \cite{SLG,ding2021street}. Compared with GCN and GAT, MPNN is good at processing different kinds of edge features (including categorical and continuous). Thus in this paper, we use MPNN as the basic GNN model for the GNN-based IP geolocation framework.%ding2021street
	
	Due to its powerful spatial modelling ability, GNN has achieved impressive performances in various graph-type data areas, such as social network \cite{zhou2022identifying}, knowledge map \cite{DBLP:journals/corr/abs-2106-08564} and recommendation system \cite{song2022deep}. Especially, there is also a kind of geolocation task in social network -- user geolocation. User geolocation (UG) is to identify the geographic location of online social network (OSN) users, such as Twitter users \cite{zhou2022identifying}. Several works such as \cite{zhou2022identifying,zhang2021heterogeneous,funes2021designing,jing2021geogat} have shown GNN's advantages in user geolocation. For example, \cite{zhou2022identifying} presents Hierarchical Graph Neural Networks (HGNN) which leverages robust signals from geographically close crowds rather than individuals. Through exploiting both unlabeled nodes and isolated nodes, HGNN achieves better performance than traditional learning-based methods. Though computer network and OSN are both represented as graph-structured data, GNN-based UG methods cannot be simply transferred to IP geolocation. First, OSNs like Twitter usually contain many kinds of node features like user posts, location tags, pictures while node features in IP geolocation are quite limited, mainly relies on network measurement data (such as delay). Second, the links between OSN users usually do not contain crucial features like delay or hop count in IP geolocation. Thus, existing GNN-based UG methods usually ignore edge features and focus on exploiting rich node features. These GNN methods may lose important information when transferred to IP geolocation tasks. GNN-based IP geolocation methods should pay attention to both node and edge features. Besides, the accuracy of existing UG methods is mainly around city-level while existing fine-grained IP geolocation methods have already achieved street-level. The newly proposed GNN-based fine-grained IP geolocation methods should also achieve street-level.
	
	Recently, researchers also begin to try GNN in different problems of computer networks \cite{WeiweiJiang2021GraphbasedDL}. For example, network modeling is used to reconstruct the computer networks and predict the structure/topology of unseen computer networks. GNN has been introduced into network modelling in \cite{badia2019towards,MiquelFerriolGalmes2020ApplyingGD}. \cite{FabienGeyer2019DeepTMAPE, FabienGeyer2021GraphBasedDL} discuss how to utilize GNN in network calculus analysis. GNN is also helpful in link delay prediction \cite{rusek2018message} and network traffic prediction \cite{TanwiMallick2020DynamicGN,ChenmingYang2020MSTNNAG,ZhaoJianlong2020SpatiotemporalGC}. GNN has been introduced into automatic detection for Botnets, which is important to prevent DDoS attacks\cite{JiaweiZhou2020AutomatingBD}. However, most previous GNN-based works in computer networks are focusing graph-level or edge-level classification or regression tasks. Little attention has been paid to how to model the relationship between location signals from network measurement data for IP geolocation (node regression/classification) until now.

    \section{IP geolocation Formulation}%all formulate change to system model
    \label{Problem_Formulation}
    To explore the potential of GNN in fine-grained IP geolocation, we need to formulate IP geolocation as a graph-based learning problem at first. \textcolor{black}{Before introducing the formulation, we want to clarify several issues here. Since there have been many mature coarse-grained IP geolocation methods, in this paper, we assume the coarse-grained locations of targets have been obtained before fine-grained geolocation. Thus, when mentioning a computer network in this work, it usually means a city-level network or at maximum state-level network. We also suggest researchers to select a most suitable coarse-grained method in each area. Since coarse-grained methods' performances may often change in different areas, and more accurate coarse-grained location results usually help a lot for following fine-grained IP geolocation activities.}

    In this paper, the IP addresses ${IP_{i}, i=1,2,...,N_{IP}}$ and the links ${LK_{j}, j= 1,2,...,N_{LK}}$ between IP addresses are used to represent the topology of a computer network. The link means a direct physical link between two IP addresses. The location of each IP address is represented as a pair of (latitude, longitude). Each IP address is associated with measured attributes, such as the delay from the probing host. Each link is also associated with measured attributes, such as the delay of the link. All IP addresses can be divided into four groups: the probing host IP address, the landmark IP addresses, the target IP addresses and the router IP addresses. The router IP addresses are the intermediate router IP addresses found by the probing host when tracerouting the landmarks and targets.

    \begin{figure}[t]
        \centering
        \includegraphics[width=3.3in]{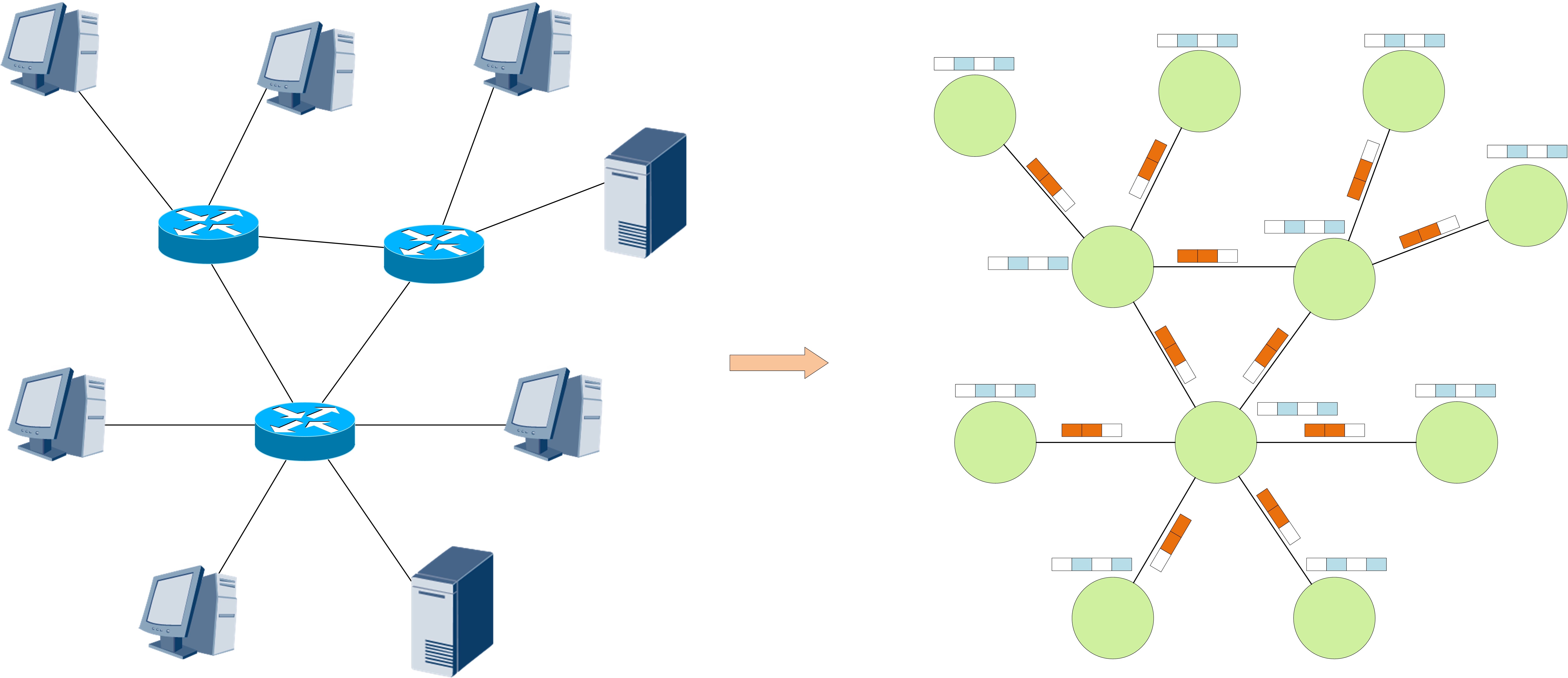}
        \label{fig:problem}
        \caption{Mapping a Computer Network into An Attributed Graph}
    \end{figure}

    \indent
    As shown in Fig. 1, we map a computer network topology to an attributed undirected graph, $G = (V, E, XV, XE)$. In the graph, a graph node $v_i \in V,i \in \{0,1,\dots,N_V\}$ represents an IP address $IP_{i}$. Each link $LK_{j}$ is represented as a graph edge $e_j \in E,j \in \{0,1,\dots,N_E\}$. Let $xv_{i}\in XV, i \in \{0,1,\dots,N_V\}$ and $xe_{j}\in XE,j \in \{0,1,\dots,N_E\}$ represent the attributes of the node $v_i$ and the edge $e_j$, respectively. The latitude and longitude location of $v_i$ are $loc_{v_{i}}$. The two dimensions of $loc_{v_{i}}$ are $[-90,90]$ and  $[-180,180]$, respectively. Assume the nodes belong to the probing host, landmarks, targets and routers are $V_p$, $V_l$, $V_t$ and $V_r$. Then the locations of the probing host and landmarks are $\{loc_{v_{i}}, {v_{i}} \in \{V_p, V_l\}\}$. The locations of targets are $\{loc_{v_{i}}, {v_{i}} \in V_t\}$. Herein, the measurement-based IP geolocation problem is formulated to learn a prediction function $f(\cdot)$, which can estimate the targets' location:
    %\ref{fig:problem}
    \begin{equation}
        [\{loc_{v_{i}}, {v_{i}} \in \{V_p, V_l\}\} ; G] \stackrel{f(\cdot)}{\rightarrow} \{\widehat{loc_{v_{i}}}, {v_{i}} \in V_t\},
    \end{equation}
    where the outputs are targets' estimated locations $\{\widehat{loc_{v_{i}}}, {v_{i}} \in V_t\}$, and the inputs are (i) the attributed undirected graph $G = (V, E, XV, XE)$; (ii) the known locations of the probing host and landmarks $\{loc_{v_{i}}, {v_{i}} \in \{V_p, V_l\}\}$.  In this paper, we use the numerical values of (latitude, longitude) to represent the location of each node, thus this is an attributed graph node regression problem. If the location is categorical, like communities or streets, the IP geolocation problem can also be formulated into a node classification problem. Note that researchers usually do not have the location information of intermediate routers $V_r$, thus this is also a typical semi-supervised problem. It is worth reporting here, it is also easy to estimate the location of intermediate routers by formulating IP geolocation into a graph node regression problem. Because the graph representations of router nodes can be naturally learned along with target nodes.%In next section, we leverage GCN to learn the representations of each node.

    %\begin{figure}[t]
%		\centering
%		\includegraphics[width=3.5in]{fig/gcn_mlp.jpg}
%		\caption{The Difference between MLP-Geo and a Possible GCN-based IP geolocation Method}
%		\label{fig:gcn_mlp}
%    \end{figure}

    \section{\textcolor{black}{THE POSSIBLE VALUE OF GNN FOR IP GEOLOCATION}}
    \label{sec:compare}
    Generally speaking, the two most important phases in learning-based IP geolocation methods are: (i) target embedding and (ii) target location mapping. From previous methods like \cite{SLG,ding2021street,zhang2020geolocation}, we know raw network measurement data contains useful signals which can help in IP geolocation. The target embedding phase is responsible to extract all useful signals from raw network measurement data, and then compress them into vectorized representations of target IP addresses (i.e., targets' embeddings). Then the target location mapping phase is responsible to map targets' embeddings to targets' locations. \textbf{The main problem of MLP-Geo is that its target embedding phase loses many useful signals from measurement data}.\\
    %\indent
    %In Fig. \ref{fig:gcn_mlp}, we show the previous learning-based method MLP-Geo and a possible GNN-based IP geolocation method.
    %We mainly focus on the differences between their target embedding phase. As shown in Fig. \ref{fig:gcn_mlp},
    \indent
    MLP-Geo consists of three dense layers \cite{zhang2020geolocation}. Here we assume its target embedding phase includes the first two layers. And its target location mapping phase is the last layer. Then the targets' embeddings $E_{2}^{mlp}$ of MLP-Geo are generated as follows:
    \begin{align}
        \label{formula:mlp_geo_embedding}
        E_{2}^{mlp} &=  f_{2} ({f}_{1} (Input)),\\
        f_{i} &= \sigma (W_{i}X_{i}+ {b}_{i}),
    \end{align}
    where $Input$ is the input features of MLP-Geo, ${f}_{i}$ denotes the $i$-th-dense-layer of MLP-Geo, ${W}_{i}$ is the weight of the $i$-th-dense-layer, and ${b}_{i}$ is the bias of the $i$-th-dense-layer. $X_{i}$ is the input of $i$-th-dense-layer and the output of the $(i-1)$-th-dense-layer. $Input$ is $X_{0}$. In MLP-Geo, $Input$ consists of the delay and router IDs from the probing host to the targets.\\
    \indent
    From Formula 3, we can see the dense layer ${f}_{i}$ uses matrix multiplication to extract useful signals from the input data. In this way, it can only judge whether a router ID shows in the path between the probing host and a target. MLP-Geo cannot extract the sequence relationship of the routers between the probing host and targets. It also cannot describe the topology around an IP addresses (or the whole network). And it cannot leverage the delay between two directly-linked IP addresses, which have been proved useful by the rule-based methods like SLG. It is hard for MLP-Geo to extract these useful signals because it is not designed for non-Euclidean structure data like graph.\\%Besides, it cannot consider the influence of the digits of IP addresses. For example, 110.220.100.32 and 110.220.100.35 may belong to one organization and are near to each other.
    \indent%  a deep learning method designed for graph-structured data
    The core concept of GNN (or more precisely, MPNN in this paper) is message passing \cite{gilmer2017neural}. As introduced in last section, all IP addresses are transformed into graph nodes. The target embedding phase of GNN is as follows:%As shown in Fig. \ref{fig:gcn_mlp},
    \begin{align}
        \label{formula:gcn_geo_embedding}
        E^{gnn}_{i} &= f_{i}(\dots(f_{2} ({f}_{1} (E_{0})),\\
        f_{i} &= \{f_{message},f_{aggregate},f_{update}\},\\
        M_{i} &=  f_{message}(E^{neighbor}_{i-1},E^{link}),\\
        MS_{i}&= f_{aggregate}(M_{i}),\\
        E^{gnn}_{i} &= f_{update}(MS_{i},E^{gnn}_{i-1}),
    \end{align}
    where $f_{i}$ indicates the $i$-th-MP-layer, $E^{gnn}_{i}$ is the node embeddings of the $i$-th-MP-layer, $E_{0}$ denotes the input (initial node embeddings). We can see both GNN and MLP learn new embeddings for IP addresses over multiple layers. GNN is a general framework. One MP layer $f_{i}$ consists of three functions: the message function $f_{message}$, the aggregate function $f_{aggregate}$ and the update function $f_{update}$. For a node, the message function $f_{message}$ passes its neighbor node's embedding ($E^{neighbor}_{i-1}$) along the edge/link ($E^{link}$) to it. A neighbor node means there is one edge between these two nodes. Then the aggregate function $f_{aggregate}$ collects the aggregation of all its neighbor nodes' messages ($MS_{i}$). The update function $f_{update}$ forms the node's new embedding $E^{gnn}_{i}$ based on the aggregated messages $MS_{i}$ and the node's previous embeddings $E^{gnn}_{i-1}$.\\
    \indent
    From Formula 4-8, we can see the embedding of each graph node also aggregates with neighbor nodes' embeddings and its edge embeddings at the beginning of each MP layer. And in $i$-th-MP-layer, a node can get information from its $i$-hop neighbors. For example, in the $2nd$-MP-layer, a node not only gets messages from its $1$-hop neighbors but also gets information from $2$-hop neighbors. Because the messages from its $2$-hop neighbors are passed to its $1$-hop neighbors in $1st$-MP-layer. Through passing messages along the edges, after several MP layers, every node will gradually know the topology and node/edge attributes of the computer network. So compared with MLP, GNN can naturally describe the network topology in the targets' embeddings. And the other features like delays of links and IP addresses can also be utilized, because they can be easily combined into message function $f_{message}$, or the embeddings of links $E^{link}$. \textcolor{black}{Thus, in some networks, if their measurement data (such as network topology, delays between routers and IP addresses) contain useful information for IP geolocation, and also is enough for GNN to learn (deep learning methods would overfit if dataset is too sparse), then GNN may show better performance in these networks. Next, to test how better GNN could be in real IP geolocation tasks, we design a GNN-based IP geolocation framework and carry out experiments to measure its performance in various real networks.}
    \begin{figure*}[t]
		\centering
        %\left
		\includegraphics[width=7.3in]{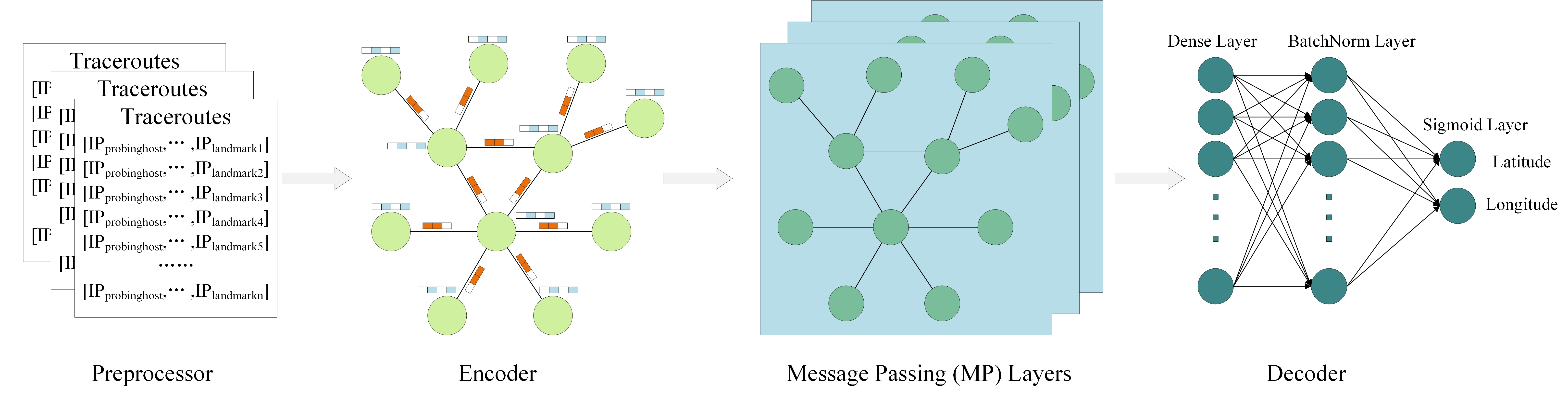}
		\caption{Illustration of the GNN-Geo Framework}
		\label{fig:model}
    \end{figure*}
    \section{Our Proposed GNN-Geo Framework}
	\label{Model}
    We present the GNN-Geo framework in this section. Fig. \ref{fig:model} illustrates the framework, which consists of four components: a preprocessor, an encoder, MP layers, and a decoder. The preprocessor maps the raw measurement data of a computer network into its graph representation $G = (V, E, XV, XE)$. The encoder generates the initial feature embeddings of $G$. The MP layers refine the initial feature embeddings with graph signals and output the refined embeddings for all nodes $V$. Finally, the decoder maps the refined node embeddings into the locations. We will describe each component one by one, followed by the optimization details. It is worth mentioning that by changing some details of each component, we can get different GNN-based IP geolocation methods from the GNN-Geo framework.

    \subsection{The Preprocessor}
    In this paper, the measurement data is mainly traceroute data from the probing host to landmarks and targets. The task of the preprocessor is transforming the raw traceroute data into the initial embeddings of $G = (V, E, XV, XE)$ for the encoder. This mainly consists of two sub-tasks: (i) building the Graph $G$ with the graph nodes $V$ and the links $E$; (ii) extracting the node attributes $XV$ and the link attributes $XE$.

    \subsubsection{Building Graph for the Measured Computer Network}
    \label{Building_Graph}
    ~~
    \textbf{Node}. As mentioned in Section \ref{Problem_Formulation}, we use the IP addresses and the links between IP addresses to represent the topology of a computer network. Thus all IP addresses in the raw traceroute data are transformed into the graph nodes. Each node $v_i$ is differentiated by a node ID. The ID is from 1 to $N_V$. $N_V$ is the number of all IP addresses found in traceroute data.\\
    \indent
    \textbf{Edge}. If the probing host finds a direct physical link between two IP addresses, then the link is transformed into a graph edge $e_j$. During the traceroute, it is possible that the probing host cannot get the IP addresses and delays of some routers, which are called "anonymous routers". There are two kinds of methods to build an edge over an "anonymous router": (i) ignore the anonymous router and build an edge between the two routers before and after the anonymous router along the routing path; (ii) map the anonymous routers to an IP address found in other re-measured routing paths. Both solutions are tested in our early experiments. The difference in their final geolocation performances is not significant. This may be due to that GNN itself is a powerful feature extractor, which is not sensitive to such differences in the input graph. However, the first solution introduces more edges in $G$, which leads to higher computation complexity and greater memory needs in the optimization phase. Thus we leverage the second solution in this paper, which is introduced as follows.\\
    \indent
    After tracerouting one IP address multiple times, we may get several different routing paths. We assume that if two routing paths are the same as each other except for the anonymous routers, then the two paths are likely to be similar. Then we can map the anonymous routers in one path to the IP addresses in other similar paths. Our matching algorithm is shown in Algorithm 1. The raw path set refers to all the raw paths found by the traceroute. The completed path set refers to the paths of which the anonymous routers have been mapped to IP as much as possible. After using Algorithm 1, we can build edges based on the completed path set. We ignore any anonymous routers which are still not mapped to IP addresses. And then an edge is built between the two routers before and after the anonymous routers along the routing path. The number of all final edges is $N_E$. It is worth mentioning here researchers can replace Algorithm 1 with other methods to check its influence on the final geolocation performance in future works.

    \begin{algorithm}
    %\label{alg:match}
    \caption{Map Anonymous Routers to IP addresses}
    \LinesNumbered
    \KwIn{Raw path set ${P_{r}}$, Completed path set ${P_{c}} = \varnothing$, Flag $flg = 1$}
    \KwOut{Completed path set ${P_{c}}$}:
    \For {each ${P_{r}^i} \in P_{r}$}{
                $flg = 1$\;
                \For {each ${P_{c}^j} \in P_{c}$}{
                    \If {the hop count of ${P_{r}^i}$ equals ${P_{c}^j}$}{
                        \If {the hop of ${P_{r}^i}$ equals ${P_{c}^j}$ except anonymous routers}{
                          \textcolor{black}{Update anonymous routers in ${P_{c}^j}$ with corresponding IP in ${P_{r}^i}$}\;
                           \textcolor{black}{Update ${P_{c}^j}$ in ${P_{c}}$ with updated ${P_{c}^j}$}\;
                           $flg = 0$\;
                           \textcolor{black}{break}\;
                        }}
                }
        \If {\textcolor{black}{$flg = 1$}}{
                    \textcolor{black}{Add ${P_{r}^i}$ into ${P_{c}}$}\;}
        \Else{
                    $flg = 1$\;
        }
    }
    \textbf{final}\;
    \textbf{return} ${P_{c}}$
    \end{algorithm}

    \subsubsection{Extracting Attributes}
    \label{Extracting_Attributes}
    ~~
    \textbf{Node Attributes}. In this work, each graph node is associated with two kinds of node attributes: node delay and node IP address. The node delay is the direct delay from the probing host to the node. For each node, the delay is repeatedly measured by the probing host many times. The minimum one is selected as the node delay because it contains minimum congestion and is closer to the true propagation delay \cite{ding2021street}. In the end, we combine the node IDs and node attributes as the initial feature of a node. $\mathbf{v_{i}}$ denotes the initial features of node $v_{i}$ and $\mathbf{V}$ denotes the initial features of all nodes.\\
    \indent
    \textbf{Edge Attributes}. Each edge can be associated with several kinds of features such as the edge delay, the head node IP address, and the tail node IP address. Here, the node nearer the probing host is called the head node of the edge, and the other one is the tail node. The edge delay is calculated by subtracting the node delay of the head node from the tail node. We keep edge delay even if it could be a negative value. Since we do not need to refine the edge representation in the MP layers, we do not need edge IDs to differentiate edges. The initial features of an edge only consist of its edge attributes. $\mathbf{e_{j}}$ denotes the initial features of edge $e_{j}$ and $\mathbf{E}$ denotes the initial features of all edges.

    \subsection{The Encoder}

    The encoder aims to form initial low dimensional embeddings of graph nodes and edges for the MP layers. The embeddings of graph nodes and edges are generated from the initial node features $\mathbf{V}$ and the initial edge features $\mathbf{E}$ from the preprocessor.\\
    \indent
    For each non-zero feature in $\mathbf{V}$ and $\mathbf{E}$, we associate it with an embedding vector. We concatenate the embeddings of a node (or edge) into a vector to describe the node (or edge). To be specific, the low dimensional embeddings of a node $v_{i}$ and an edge \textcolor{black}{$e_{u}$} are:
    \begin{align}
    \label{formula:encoder}
            \mathbf{h}_{v_{i}}^{(0)}&=  \mathbf{Q_{v}}^{T}  \mathbf{v_{i}},\\
            \mathbf{h}_{e_{j}}      &=  \mathbf{Q_{e}}^{T}  \textcolor{black}{\mathbf{e_{u}}},
    \end{align}%
    where $\mathbf{Q_{v}} \in \mathbb{R}^{N_V \times G}$ denotes the embedding matrix for all nodes $V$, $\mathbf{Q_{e}} \in \mathbb{R}^{N_E \times K}$ denotes the embedding matrix for all edges $E$. $N_V$ denotes the number of node features and $G$ denotes the embedding size. $N_E$ denotes the number of edge features and $K$ denotes the embedding size.%\textcolor{black}{\sout{The encoder then feeds $\mathbf{h}_{V}^{(0)}$ and $\mathbf{h}_{E}$ into the next component -- the MP layers for embedding augmentation. It is worth reporting that pooling mechanisms could be applied here instead of concatenating, like average pooling, max pooling, attention-based pooling \cite{xin2019cfm}. However, average pooling and max-pooling tried in our early experiments did not improve the performances. This may be because they lost some information useful for geolocation. The attention-based pooling requires more parameters which could lead to overfitting and thus is neglected in this work. In the end, we decide to leverage concatenating to generate the feature vector.}}

    \subsection{Message Passing (MP) Layers}

    \textcolor{black}{The inputs of Message Passing (MP) layers are $\mathbf{h}_V^{(0)}$ and $\mathbf{h}_E$ from the encoder. MP layers will augment the initial graph node embeddings ($\mathbf{h}_V^{(0)}$) by explicitly modeling the edges between nodes and graph/edge attributes. Each MP layer consists of message, aggregate and update functions as follows.}

    \textbf{Message Function}. For a node $v_i$ and one of its neighbour nodes $v_j$, the message from $v_j$ to $v_i$ is defined as:
    \begin{align}
        \mathbf{m}_{i \leftarrow j}=f_{message}(\mathbf{h}_{v_{j}}, \textcolor{black}{\mathbf{e}_{i \leftarrow j}}),
    \end{align}
    The inputs of $f_{message}$ are: (i) the embedding of the neighbour node $\mathbf{h}_{v_{j}}$;  (ii) the embedding of the edge \textcolor{black}{$\mathbf{e}_{i \leftarrow j}$ from $v_j$ to $v_i$}. In this paper, $f_{message}$ is implemented as:
    \begin{align}
     \textcolor{black}{\mathbf{m}_{i \leftarrow j}}=f_{edge}(\textcolor{black}{\mathbf{h}_{e_{i \leftarrow j}}}) \mathbf{h}_{v_{j}},
    \end{align}
    where the edge network $f_{edge}$ is a two-layer MLP, which transforms the edge embedding \textcolor{black}{$\mathbf{h}_{e_{i \leftarrow j}}$} to a matrix \textcolor{black}{$W_{e_{i \leftarrow j}}$} for a neighbor node's message $h_{v_j}$. The architecture of $f_{edge}$ is:
    \begin{equation}
        \label{formula:our_prediction_layer}
        \textcolor{black}{\mathbf{W}_{e_{i \leftarrow j}}}= \mathbf{W}_{edge}^{2}(\operatorname{Relu}( \mathbf{W}_{edge}^{1}(\textcolor{black}{e_{i \leftarrow j}}))),
    \end{equation}
    where \textcolor{black}{$\mathbf{W}_{e_{i \leftarrow j}}$} is used as a weight to control the influence of the neighbor's message $\mathbf{h}_{v_{j}}$ to node $i$. $\mathbf{W}_{edge}^{1}$ and $\mathbf{W}_{edge}^2$ are the weight matrices of the two MLP layers. They first transform the initial edge embedding \textcolor{black}{$\mathbf{h}_{e_{i \leftarrow j}}$} (size $K$) to a temporary embedding (size $2K$), then transform the temporary embedding into a one-dimensional embedding. The one-dimensional weight embedding will be reshaped as a two-dimensional square weight matrix \textcolor{black}{$\mathbf{W}_{e_{i \leftarrow j}}$} (size $G \times G$) to control the influence of neighbor node $v_j$ to node $v_i$. The activation function of ReLU \cite{agarap2018deep} is leveraged to catch the non-linear relationships.

    The edge embedding \textcolor{black}{$\mathbf{h}_{e_{i \leftarrow j}}$} is generated from the edge attributes like the edge delay and the IP addresses of two edge nodes. $\mathbf{h}_{v_{j}}$ contains the node ID and attributes (e.g., IP addresses, the delay to the probing host). It is reasonable to use the transformed edge embeddings as message weight. For example, if the edge delay is too large, it may reflect that $v_j$ is too far away, then its message is not so important to $v_i$; Or if the two IP addresses are very similar, it may reflect that the two nodes belong to one organization, then $v_i$ should pay more attention to $v_j$. In this way, node $v_i$ will learn what is $v_j$'s information ($\mathbf{h}_{v_{j}}$), and can decide how much information it want to receive from the relationship ($f_{edge}(\textcolor{black}{\mathbf{h}_{e_{i \leftarrow j}}}$)).

    \textbf{Aggregation and Update Function}. $U_{v_{i}}$ denotes the set of neighbor nodes of $v_i$. The aggregation function $f_{aggregate}$ collects all the messages propagated from $U_{v_{i}}$ to $v_{i}$. The update function $f_{update}$ uses the collected messages $A_{v_{i}}$ to update the embedding of $v_{i}$. The aggregation and update function for $v_{i}$ are defined as:
    \begin{align}
        A_{v_{i}} &= f_{{aggregate}_{v_j \in U_{v_{i}}}}(\mathbf{m}_{i \leftarrow j}),\\
        \mathbf{h}_{v_{i}}^{(l)}&= f_{update}(\mathbf{h}_{v_{i}}^{(l-1)},A_{v_{i}}),
    \end{align}
    where $\mathbf{h}_{v_{i}}^{(l)}$ denotes the updated embedding of node $v_{i}$ after the $l$-th MP layer ($l=1,2,3,4,\dots$). $\mathbf{h}_{v_{i}}^{(0)}$ is the initial node embedding from the encoder, as shown in Formula \ref{formula:encoder}. The aggregation function $f_{aggregate}$ can be a simply symmetric function such as Mean \cite{1st_GCN}, Max \cite{hamilton2017inductive}, or Sum \cite{xu2018powerful}. Since no previous work has tested them in IP geolocation, we tried all three kinds of aggregators in experiments. The update function is implemented as follows:

    \begin{equation}
    \label{formula:update}
        \begin{aligned}
            \mathbf{h}_{v_{i}}^{(l)}= \operatorname{ReLU}({\mathbf{h}_{v_{i}}^{(l-1)} + A_{v_{i}}}),
        \end{aligned}
    \end{equation}
    where we update the node $v_i$ by summing up its previous embedding $\mathbf{h}_{v_{i}}^{(l-1)}$ and its aggregated messages from neighbors $A_{v_{i}}$. \textcolor{black}{Relu is used to do non-linear feature transformation since nodes and edges have rich semantic attributes in this paper}. In this way, the new node embedding $\mathbf{h}_{v_{i}}^{(l)}$ contains following information : (i) which nodes are its neighbors (topology nearby); (ii) its neighbors' attributes; (iii) its edges' attributes (e.g., IP address, delay); (iv) its own attributes (e.g., ID, IP address, delay). \textcolor{black}{By stacking $L$ MP layers, a node is capable of receiving the messages propagated from its $L$-hop neighbors. The final embeddings for all node $\mathbf{h}_{V}^{(L)}$ are then fed into the decoder for location estimation.}
    \subsection{The Decoder}
    \label{sec:model_decoder}
    The decoder aims to estimate the locations from the nodes' refined embeddings $\mathbf{h}_{V}^{(L)}$. We need to predict two numerical values (latitude and longitude) of all nodes. For problems in other research areas, a widely-used original decoder after MP layers consists of dense layers. For example, a typical two-dense-layers decoder is:

    \begin{equation}
        \begin{array}{c}
        \label{formula:typical_dense_layers_for_prediction}
        \widehat{loc}= \sigma  \left ( \mathbf{W}_{loc}^{2}  ( \mathbf{W}_{loc}^{1} \mathbf{h}_{V}^{(L)}+ {b}_{loc}^{1} ) +  {b}_{loc}^{2}  \right ),
        \end{array}
    \end{equation}
    where $\widehat{loc} \in \mathbb{R}^{N_V \times 2}$ denotes the estimated location matrix for all nodes $V$. The two dimensions of $\hat{loc}$ are the estimated numerical values of latitude and longitude for all nodes. $\mathbf{W}_{loc}^{1}$ and $\mathbf{W}_{loc}^{2}$ are the weights of the two dense layers. ${b}_{loc}^{1}$ and ${b}_{loc}^{2}$ are biases. $\sigma$ is the non-linear activation function to endorse nonlinearity, such as Relu in Formula \ref{formula:update}. Then we can train GNN-Geo by comparing the ground-truth locations ${loc}_{train}$ and the estimated locations $\hat{loc}_{train}$. ${loc}_{train}$ (or $\hat{loc}_{train}$) means the real/estimated locations of the training dataset.\\% We can also stack more dense layers to increase the learning ability for the decoder.
    \indent
    However, this kind of decoder is not so suitable for IP geolocation: the output can be any data in $\mathbb{R}$ while the latitude and longitude range are (90S\textasciitilde 90N, 180W\textasciitilde 180E). Thus most output data is useless and significantly increases the difficulty of optimization. Actually, even finding a fine-grained location of targets among the whole earth is not necessary. As mentioned in Section 1, fine-grained IP geolocation is on the basis of coarse-grained geolocation. Thus, we usually already know a rough location of the target before estimating its fine-grained location. Inspired by this, we \textcolor{black}{combine rules into the pure MLP-based decoder to} improve the performance of GNN-based IP geolocation methods. First, we need to use two min-max scalers to transform the ${loc}_{train}$ (latitude \& longitude) into $loc_{train}^t$. The scales of transformed latitude and longitude in $loc_{train}^t$ are both $[0,1]$. Then our decoder is implemented as follows:%

    \begin{equation}
        \label{formula:our_prediction_layer}
        \widehat{loc^{ts}}= \operatorname{Sigmoid}( \operatorname{BatchNorm}(\mathbf{W}_{loc} \mathbf{h}_{V}^{(L)} + \mathbf{b}_{loc})),
    \end{equation}
    where $\widehat{loc^{ts}}$ is the estimated transformed location matrix for all nodes $V$.  $\operatorname{Sigmoid}$ is an activation function of which output is $[0,1]$. $\operatorname{Sigmoid}$ could become saturated if input values are too large, which makes learning even harder. $\operatorname{BatchNorm}$ refers to Batch Normalization \cite{ioffe2015batch}. Batch Normalization is applied here to ease the saturating problem of Sigmoid as well as preventing overfitting. Then we train GNN-Geo by comparing ${loc}_{train}^{ts}$ and $\widehat{loc_{train}^{ts}}$. After training, $\widehat{loc}$ can be scaled from $\widehat{loc^{ts}}$ by reversely using two min-max scalers.

         \begin{figure*}[t]
        \centering
        \subfigure[New York State]{
        \begin{minipage}[t]{0.3\linewidth}
        \centering
        \includegraphics[width=2.2in,height=1.7in]{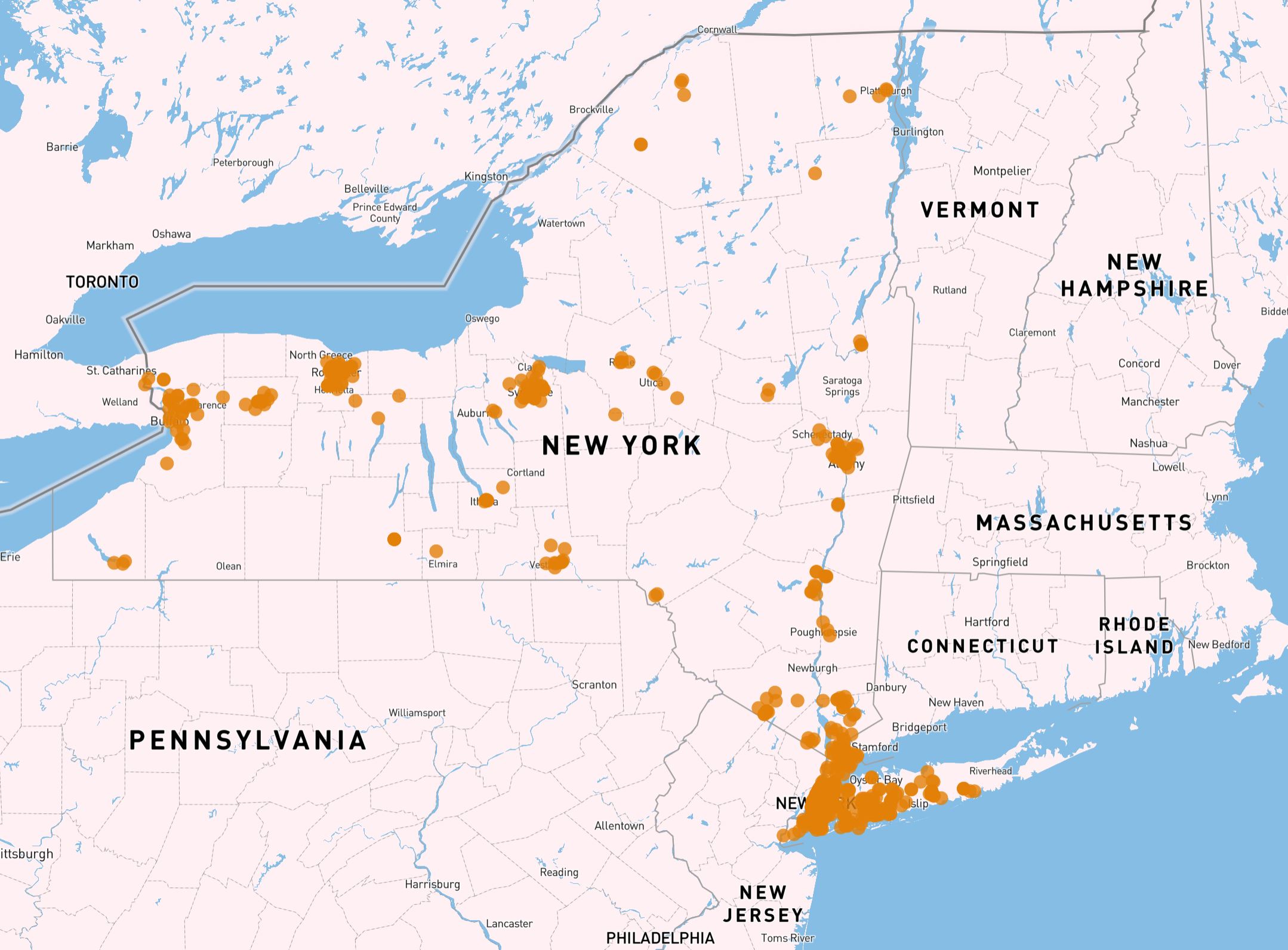}
        %\caption{fig1}
        \end{minipage}%
        \label{fig:1}
        }
        \subfigure[Hong Kong]{
        \begin{minipage}[t]{0.3\linewidth}
        \centering
        \includegraphics[width=2.2in,height=1.7in]{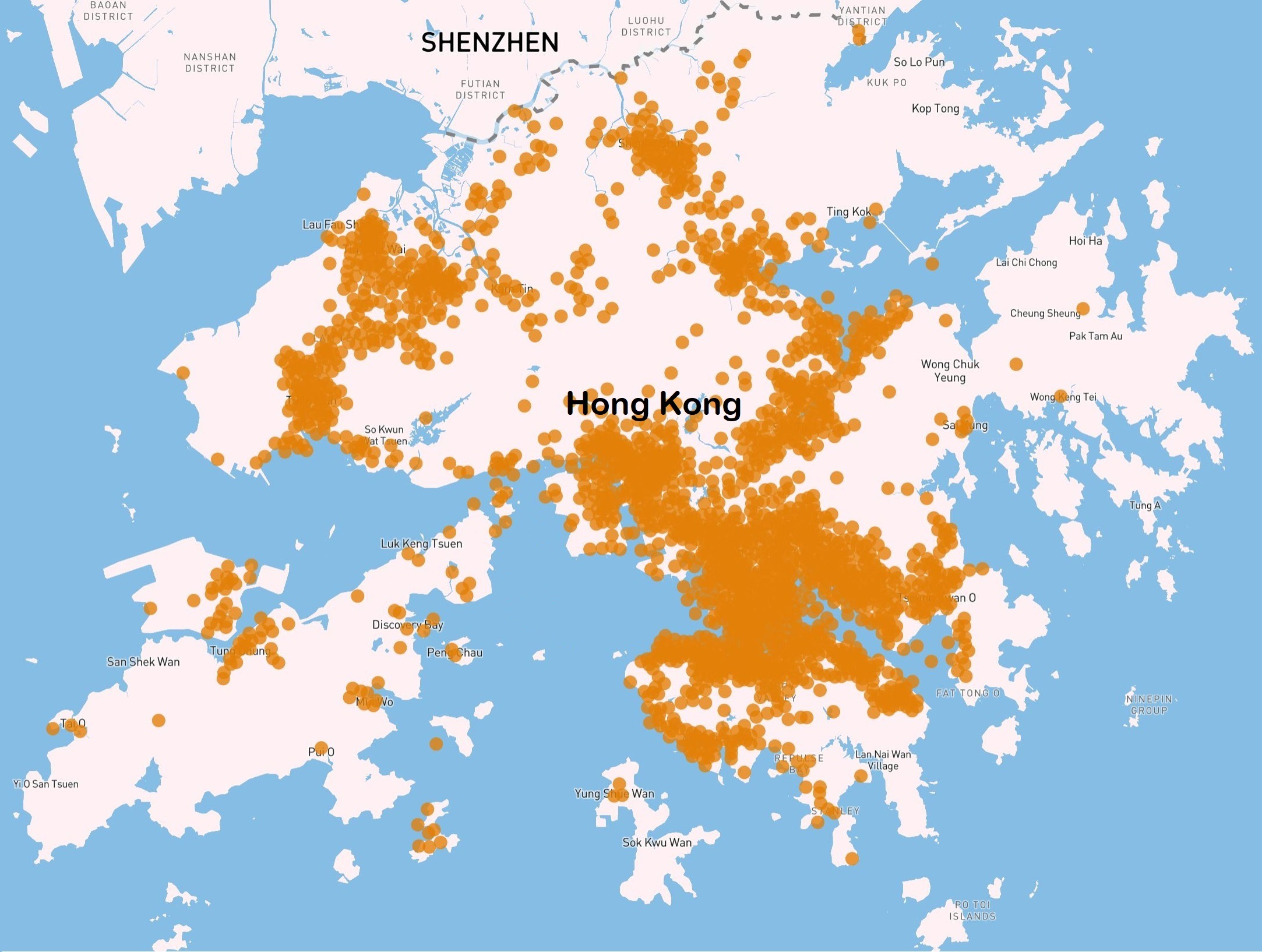}
        %\caption{fig2}
        \end{minipage}%
        \label{fig:2}
        }
        \subfigure[Shanghai]{
        \begin{minipage}[t]{0.3\linewidth}
        \centering
        \includegraphics[width=2.2in,height=1.7in]{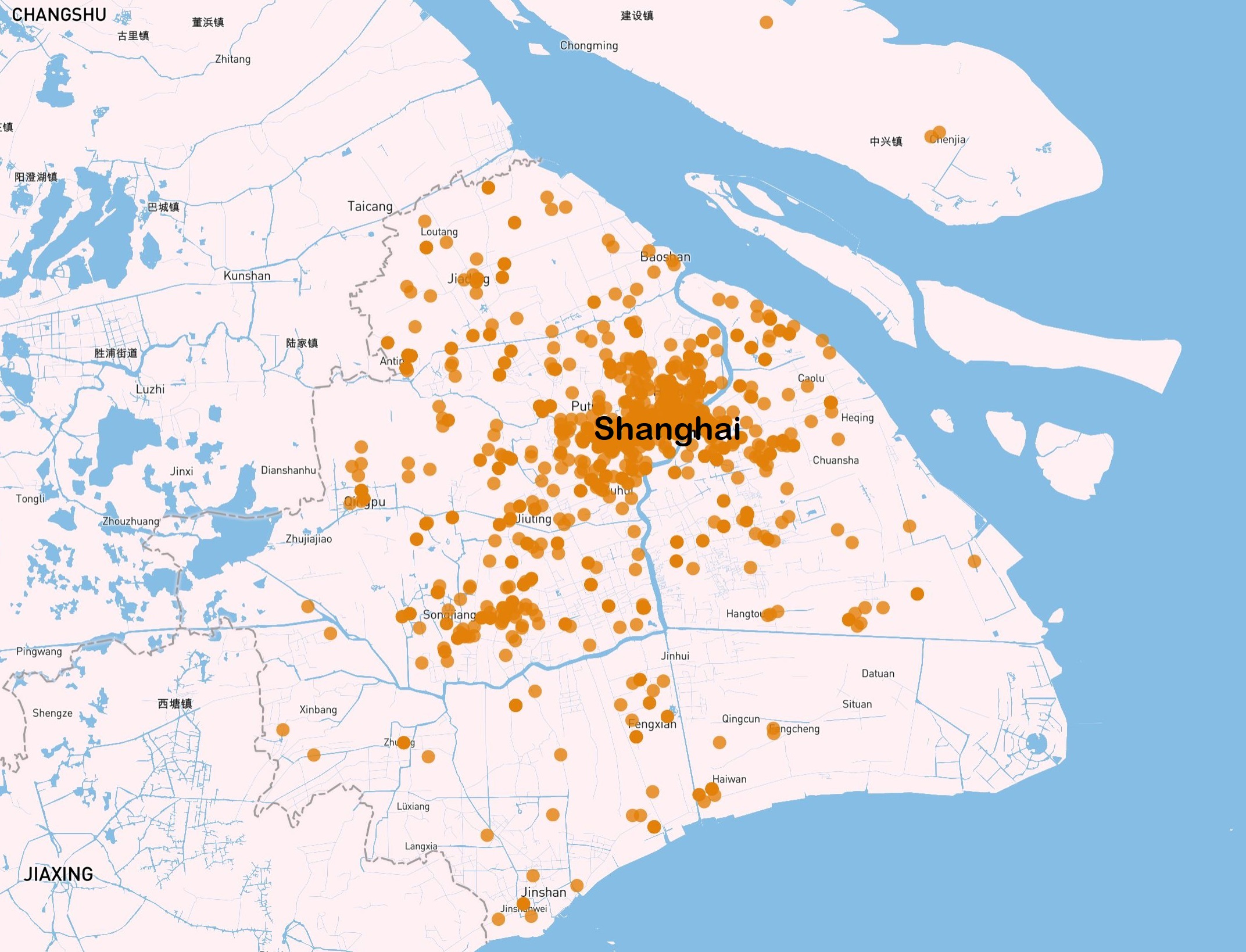}
        %\caption{fig2}
        \end{minipage}%
        \label{fig:3}
        }
        \centering
        \caption{Landmark Distribution of Three Areas as an example}
        \label{fig:landmark_map}
    \end{figure*}

    It is worth mentioning that we do not always need to scale the range of latitude and longitude of the training sets. The core idea is to find a rough area of targets, then re-scale the range of latitude and longitude of these areas to help the optimizer to converge. In this way, the decoder is actually based on a hypothetical rule that we could get a coarse-grained location of targets before fine-grained geolocation. This hypothesis is common to previous fine-grained methods. For example, SLG needs to estimate the coarse-grained location before estimating the fine-grained location of a target. And since SLG maps target IP addresses to landmarks, the largest estimated area is actually limited to the area that the known landmarks can cover. MLP-Geo also needs to cluster the landmarks and targets at first based on their measurement data, which is similar to coarse-grained IP geolocation. Corr-SLG also assumes that researchers already knew the city location of the targets and focus on finding fine-grained locations inside the city. This \textcolor{black}{improved} decoder can ease the burden of fine-tuning parameters by introducing some prior knowledge into GNN-based methods.

    \subsection{Model Training}
    To learn model parameters, we employ the Mean Square Error (MSE) loss between ${loc}_{train}$ and $\hat{loc}_{train}$. In this work, we optimize the $L_{2}$ regularized MSE loss as follows:

    \begin{equation}
    \label{formula:mseloss}
        \mathbf{Loss}=\sum_{i=1}^{N(v)_{train}} ({loc}_{train}^{ts} - \widehat{loc}_{train}^{ts})^{2}+ \lambda\|\Theta\|^{2} ,
    \end{equation}
    where $N(v)_{train}$ denotes all nodes in the training dataset.  And $\mathbf{\Theta}$ includes all the trainable model parameters of GNN-Geo. Deep learning methods usually suffer from overfitting. Besides Batch Normalization in the decoder, we also leverage $L_2$ regularization to prevent overfitting. Note that Batch Normalization is only used in training, and must be disabled during testing. $\lambda$ controls the $L_{2}$ regularization strength. The Adam optimizer is employed to optimize the prediction model and update the model parameters by using the gradients of the loss function.

    \section{Performance Analysis}
    \label{sec:evaulation}

    We evaluate experiments on \textcolor{black}{8} real-world datasets, aiming to answer the following research questions:
    \begin{itemize}
    \item RQ1: how does GNN-Geo perform as compared with state-of-the-art rule-based and learning-based street-level IP geolocation methods?
    \item RQ2: \textcolor{black}{how does diffferent training-testing ratio affect GNN-Geo's performance?}
    \item RQ3: \textcolor{black}{how can rules help to improve GNN-Geo's vanilla MLP-based decoder?}
    \item RQ4: how do different settings (e.g., \textcolor{black}{basic GNN models,} $f_{aggregation}$) affect the geolocating performance of GNN-Geo?
    \end{itemize}

    \subsection{Dataset Description}
    \label{sec:eva_dataset}
    To demonstrate the effectiveness and generalization capabilities of GNN-Geo, we plan to conduct experiments in various real-world computer networks. Due to privacy or intellectual property concerns, the datasets of most previous related works are not opened. So we need to collect our own datasets.  \\
    \indent
    \textbf{Target Areas and Landmarks}. To test the generalization capabilities of GNN-Geo in different network environments, we collect \textcolor{black}{different landmark datasets from 8 real-networks: (i) the IPv4 networks of New York State, Hong Kong, Shanghai, Beijing, and Tokyo; (ii) the IPv6 networks of Tokyo, Berlin and Munich. In section 6, they are referred as NYS, HK, Shanghai, Beijing, Tokyo, Tokyo-IPv6, Berlin-IPv6 and Munich-IPv6 datasets, respectively. These networks belong to three continents (North America, Asia and Europe), including both developed countries and developing countries. The internet service providers (ISP) in these areas are also different. For example, the main ISPs are Spectrum and Verizon in NYS; Asia Netcom in HK; China Telecom and China Unicom in Shanghai and Beijing. The urban environment, and terrain of these areas are also different. For example, NYS consists of several cities, including New York City, Buffalo city, Rochester city, etc. HK mainly consists of Hong Kong Island, Kowloon Peninsula, New Territories, and some smaller islands. Shanghai consists of the main urban area on the Changjiang alluvial plain as well as rural areas on a large island called Chongming Island. All 8 networks cover densely populated urban areas and sparsely populated rural areas. The sizes of these areas are also different as shown in Table 1}.\\
    \indent
    \textcolor{black}{We collect fine-grained IPv4 landmarks following the method in [5], which has been widely used by previous works. These landmarks are mainly famous organization website servers, whose IP addresses and organization locations are opened to the public. However, since more and more new website servers are deployed in the clouds, this collection method is not suitable for finding enough IPv6 fine-grained landmarks. Recently, Erik Rye and Robert Beverly \cite{rye2022ipvseeyou} find many WiFi devices' IPv6 addresses and their geographical locations can be connected by their MAC address and WiFi BSSID (Basic Service Set Identifier). Based on this method, we collect IPv6 landmarks all over the world. In total, we find more than 350,000 IPv4/IPv6 landmarks.  After verification and selection, there are 11,654 measurable fine-grained landmarks. Each landmark is an IP address with a pair of (latitude, longitude). Due to limited space, we only show the landmark distribution in NYS, HK, and Shanghai in Fig. 3. The covering area and density of landmarks also could affect the performance of IP geolocation methods \cite{ciavarrini2018geolocation}. The number of landmarks in each network are shown in Table 1}.

    \subsection{Engineering Applications of the Preprocessor and the Encoder}
    We deploy one probing host in \textcolor{black}{each target network. The probing host is IPv4 (or IPv6) if the target network is IPv4-based (IPv6-based)}. Every probing host is responsible for getting the raw traceroute data in the area where it locates. The network graph of the area is then built based on the measurement data obtained by its probing host. For 10 days, one probing host probes all landmarks in its city every 10 minutes (1440 times in total). Each probing host is a server installed with network probing tools, of which function is similar to traceroute. The software can get the IP addresses of routers between the probing host and all landmarks. And it can also obtain the delays (accurate to milliseconds) between each router and the probing host. After the preprocessor, the number of nodes (IP addresses) and the number of the links are (2872, 3303) in NYS, (3071, 3697) in HK, (2530, 15108) in Shanghai, \textcolor{black}{(3310, 18742) in Beijing, (1498, 4408) in Tokyo, (2297, 4966) in Tokyo-IPv6, (2070, 5776) in Berlin-IPv6 and (380, 856) in Munich-IPv6. So, their edge/node ratio is 1.15, 1.20, 5.97, 5.66, 2.94, 2.16, 2.79 and 2.25, respectively}.\\
    \indent
    The initial attribute feature for each node is a 15-dimensional vector of values. The first 5 dimensions are the raw numerical of node delay and IP address. The numerical node delay is then binned into 10 intervals to form the last 10 dimensions. Here we use K-means to cluster delay into 10 bins. 10 is empirically selected. IP addresses can also be transformed into categorical features. The initial attribute feature for each edge is an 11-dimensional vector of values. The first dimension is the raw numerical data of the edge delay. The remaining 10 dimensions are the binned edge delay.\\
    \indent
    Actually, it is easy to combine more kinds of features into the proposed GNN-Geo, such as the domain name of routers, the two IP addresses of an edge, the delay changing trends of a node or edge, etc. All these features can be transformed into node/edge attribute features by the preprocessor, and then naturally introduced into model training through the encoder. And efforts can also be put into more sophisticated pre-processing techniques. For example, we can observe the histogram distribution of delay, and manually decide the number of bins for each area instead of using empirical numbers. IP addresses can also be explicitly transformed into categorical features. However, these features or preprocessing techniques are not easy to be applied in previous baselines for comparison. In this work, we focus on testing the learning potential of GNN in IP geolocation. Thus we utilize a simple preprocessor in this work, leaving further investigation on the preprocessor to future research.

    \subsection{Experiments Settings}
    \textbf{Evaluation Metrics}. In each \textcolor{black}{network, we use 4 kinds of training/validation/testing ratios to compare the performance of GNN-Geo with baselines: training-70\%, training-50\%, training-30\%, and training-10\%. For all ratios, the validation ratio is 20\%. We mainly change the ratio of training dataset and testing dataset: for training-70\%, 70\% landmarks are randomly chosen for training, 20\% for validation, and 10\% for testing; for training-50\%, 50\% landmarks are randomly chosen for training, 20\% for validation, and 30\% for testing; for training-30\%, 30\% landmarks are randomly chosen for training, 20\% for validation, and 50\% for testing; for training-10\%, only 10\% landmarks are randomly chosen for training, 20\% for validation, and 70\% for testing. When answering RQ1 (Section 6.5 and 6.6), RQ3 (Section 6.8) and RQ4 (Section 6.9), we mainly use train-70\% datasets. Training-50\%, training-30\%, and training-10\% are mainly used in answering RQ2 (Section 6.7)}.

    We use the validation sets to do early stopping and hyper-parameters fine-tuning. We compare performances on the test sets if not specially explained. For each training set, the latitude and longitude ranges of landmarks are extended by 0.1 decimal degree, before being scaled to the $[0,1]$. For example, assume the latitude and longitude range of one training set is (30.61N \textasciitilde\ 31.73N, 121.14E \textasciitilde\ 121.86E), then the range for scaling is (30.51N \textasciitilde\ 31.83N, 121.04E \textasciitilde\ 121.96E). 0.1 decimal degree is an empirical number, about 11.1 km for both longitude and latitude. We minimize the mean squared error loss following Formula \ref{formula:mseloss}. During the validation and test phase, we need to re-scale the output of the decoder into latitude and longitude values to calculate the geographical distance. The error distance of an IP address is the geographical distance between the ground-truth location and the estimated location. To evaluate the performance of IP geolocation, we adopt \textcolor{black}{2} evaluation metrics widely-used in previous works like \cite{SLG,ding2021street}: average distance error, \textcolor{black}{and} median distance error. We also leverage Cumulative Distribution Function (CDF) to show the distribution of error distances.\\%pros cons
    \indent
    \textbf{Parameter Settings}. We implement GNN-Geo in Pytorch. Code and hashed datasets will be opened upon acceptance. The embedding sizes of the initial node ID feature in the encoder and the hidden node representation in the MP layers are searched between $\{32, 64, 128, 256\}$. We optimize the model with the Adam optimizer. We leverage the default initializer of Pytorch to initialize the model parameters and embeddings. For example, standard normal distribution $N(0,1)$ is used to initialize node ID embeddings. A grid search is applied to find the best hyper-parameters \cite{liashchynskyi2019grid}. The learning rate is tuned amongst $\{0.01, 0.001\}$, the coefficient of $L_{2}$ regularization is searched in $\{0.01,0.001,0.0005\}$. The depth of the MP layers $L$ is searched in $\{1,2,3,4,5\}$. The number of hidden units of the edge network is searched in $\{4,8,16\}$. The dimension of initial edge embedding $K$ is searched between $\{16, 8, 4\}$. The aggregation methods are searched among Mean, Sum and Max. Besides, the early stopping strategy is performed: training is stopped if the average distance error on the validation set does not increase for 1000 successive epochs. Usually, 4000 epochs are sufficient for GNN-Geo to converge.

    \begin{table*}[]
\centering
\renewcommand\arraystretch{1.2}
\caption{\textcolor{black}{Performance (kilometers) Comparison of baselines and GNN-Geo}}
\begin{threeparttable}
\begin{tabular}{c|c|c|c|cc|cc|c}
\hline
\multirow{2}{*}{Areas}          & \multirow{2}{*}{\#Landmarks} & \multirow{2}{*}{Size($km^2$)} & \multirow{2}{*}{\begin{tabular}[c]{@{}c@{}}Error \\ Distance\end{tabular}} & \multicolumn{2}{c|}{Rule-based}                     & \multicolumn{2}{c|}{Learning-based}                & \multirow{2}{*}{\begin{tabular}[c]{@{}c@{}}\%Error \\ Reduced\end{tabular}} \\ \cline{5-8}
                                &                              &                                              &                                                                            & \multicolumn{1}{c|}{SLG \cite{SLG}}          & Corr-SLG \cite{ding2021street}        & \multicolumn{1}{c|}{MLP-Geo \cite{zhang2020geolocation}}     & GNN-Geo         &                                                                             \\ \hline
\multirow{2}{*}{New York State} & \multirow{2}{*}{1705}        & \multirow{2}{*}{141299}                     & Average                                                                    & \multicolumn{1}{c|}{38.176}       & {\ul 38.303}    & \multicolumn{1}{c|}{43.274}      & \textbf{27.198} & 28.76\%                                                                     \\ \cline{4-9}
                                &                              &                                              & Median                                                                     & \multicolumn{1}{c|}{7.090}        & {\ul 7.040}     & \multicolumn{1}{c|}{13.847}      & \textbf{4.901}  & 30.38\%                                                                     \\ \hline
\multirow{2}{*}{Hong Kong}      & \multirow{2}{*}{2061}        & \multirow{2}{*}{1108}                       & Average                                                                    & \multicolumn{1}{c|}{13.019}       & 10.001          & \multicolumn{1}{c|}{{\ul 9.793}} & \textbf{8.166}  & 16.61\%                                                                     \\ \cline{4-9}
                                &                              &                                              & Median                                                                     & \multicolumn{1}{c|}{11.933}       & {\ul 7.523}     & \multicolumn{1}{c|}{9.140}       & \textbf{6.596}  & 12.32\%                                                                     \\ \hline
\multirow{2}{*}{Shanghai}       & \multirow{2}{*}{1387}        & \multirow{2}{*}{6340}                       & Average                                                                    & \multicolumn{1}{c|}{14.245}       & {\ul 14.037}    & \multicolumn{1}{c|}{14.291}      & \textbf{10.368} & 26.14\%                                                                     \\ \cline{4-9}
                                &                              &                                              & Median                                                                     & \multicolumn{1}{c|}{{\ul 11.056}} & \textit{11.281} & \multicolumn{1}{c|}{11.659}      & \textbf{8.698}  & 21.33\%                                                                     \\ \hline
\multirow{2}{*}{Beijing}        & \multirow{2}{*}{1835}        & \multirow{2}{*}{16410}                         & Average                                                                    & \multicolumn{1}{c|}{{\ul 12.394}} & 13.855          & \multicolumn{1}{c|}{12.992}      & \textbf{9.523}  & 23.16\%                                                                     \\ \cline{4-9}
                                &                              &                                              & Median                                                                     & \multicolumn{1}{c|}{{\ul 10.259}} & 11.314          & \multicolumn{1}{c|}{10.985}      & \textbf{7.603}  & 25.89\%                                                                     \\ \hline
\multirow{2}{*}{Tokyo}          & \multirow{2}{*}{943}         & \multirow{2}{*}{14034}                       & Average                                                                    & \multicolumn{1}{c|}{15.321}       & {\ul 12.978}    & \multicolumn{1}{c|}{13.047}      & \textbf{12.306} & 5.18\%                                                                      \\ \cline{4-9}
                                &                              &                                              & Median                                                                     & \multicolumn{1}{c|}{13.206}       & 10.217          & \multicolumn{1}{c|}{{\ul 9.701}} & \textbf{8.654}  & 10.79\%                                                                     \\ \hline
\multirow{2}{*}{Tokyo (IPv6)}   & \multirow{2}{*}{1949}        & \multirow{2}{*}{14034}                       & Average                                                                    & \multicolumn{1}{c|}{15.372}       & 7.066           & \multicolumn{1}{c|}{{\ul 3.912}} & \textbf{2.916}  & 25.46\%                                                                     \\ \cline{4-9}
                                &                              &                                              & Median                                                                     & \multicolumn{1}{c|}{10.813}       & 5.702           & \multicolumn{1}{c|}{{\ul 1.806}} & \textbf{1.344}  & 25.58\%                                                                     \\ \hline
\multirow{2}{*}{Berlin (IPv6)}  & \multirow{2}{*}{1503}        & \multirow{2}{*}{883}                         & Average                                                                    & \multicolumn{1}{c|}{7.214}        & {\ul 6.322}     & \multicolumn{1}{c|}{7.294}       & \textbf{3.368}  & 46.73\%                                                                     \\ \cline{4-9}
                                &                              &                                              & Median                                                                     & \multicolumn{1}{c|}{5.141}        & {\ul 5.100}     & \multicolumn{1}{c|}{6.300}       & \textbf{3.230}  & 36.67\%                                                                     \\ \hline
\multirow{2}{*}{Munich (IPv6)}  & \multirow{2}{*}{271}         & \multirow{2}{*}{310}                         & Average                                                                    & \multicolumn{1}{c|}{7.908}        & 5.028           & \multicolumn{1}{c|}{{\ul 4.270}} & \textbf{3.875}  & 9.25\%                                                                      \\ \cline{4-9}
                                &                              &                                              & Median                                                                     & \multicolumn{1}{c|}{7.890}        & 4.312           & \multicolumn{1}{c|}{{\ul 3.416}} & \textbf{3.230}  & 5.44\%                                                                      \\ \hline
\end{tabular}
        \begin{tablenotes}
            \item[1] {\ul underline} indicates the best metrics among baselines.
        \end{tablenotes}
        \end{threeparttable}
        \label{tab:mainresults}
\end{table*}

    \subsection{Baselines}
    \label{bs_line}
    To evaluate the performance of GNN-Geo, we compare it with several state-of-the-art rule-based and learning-based street-level IP geolocation methods as follows.\\
    \indent
    \textbf{SLG} \cite{SLG}. SLG is a typical rule-based IP geolocation method and is also one of the most widely-used street-level IP geolocation baselines. Its core rule is to map a target IP to the landmark which has the smallest relative delay to the target IP. There is no need to set or tune hyper-parameters for SLG. Thus in each area, we combine the training set and validation set to geolocate the testing set.\\
    \indent
    \textbf{Corr-SLG} \cite{ding2021street}: Corr-SLG is a recent rule-based IP geolocation method. It divides landmarks into 1) Group A, strong-positive delay-distance-correlated landmarks; 2) Group B, strong-negative delay-distance-correlated landmarks; 3) Group C, weak delay-distance-correlated landmarks. The main hyper-parameters are $C_a$ and $C_b$ in Corr-SLG. They are used to divide all landmarks into Group A, Group B, and Group C based on delay-distance correlation. $C_a$ and $C_b$ are manually tuned among $\{0,0.1,0.2,\dots,0.9\}$ and $\{0.1,-0.2,\dots,-0.9,-1\}$, respectively. We split the landmarks into a training set, a validation set and a testing set (same to GNN-Geo). In each area, we use the validation set to tune $C_a$ and $C_b$ , and then report the results on the test set. The parameter searching methods are the same as \cite{ding2021street}.\\
    \indent
    \textbf{MLP-Geo} \cite{zhang2020geolocation}. MLP-Geo is a learning-based IP geolocation method. Like GNN-Geo and Corr-SLG, we use validation sets to tune the hyperparameters of MLP-Geo in each area. The main hyper-parameters of MLP-Geo include the learning rate, the number of training epochs and the dimension size of the middle dense layer. A grid search is applied to find the best hyper-parameters. The learning rate is tuned amongst $\{0.1, 0.01, 0.001, 0.0001\}$. The dimension size is searched in $\{32, 64, 128, 256\}$. The number of training epochs is 20,000 to find the epoch corresponding to the best performance. The other settings are kept the same as suggested in \cite{zhang2020geolocation}. For example, we set its $beta$ to 30 when encoding the path.

    \subsection{Geolocation Performance Comparison between GNN-Geo and Baselines}
    \label{sec:results_comparison}
    The results of \textcolor{black}{all methods on 8 datasets are shown in Table 1. The datasets are train-70\% datasets}. The numbers in Table \ref{tab:mainresults} are averaged by 5 times of train-validation-testing (randomly changing seeds when splitting datasets). To achieve the best performance of all methods, parameter tuning of all methods is conducted as introduced in Section \ref{bs_line}. From Table \ref{tab:mainresults}, \textbf{we can see that the proposed framework GNN-Geo consistently outperforms all baselines w.r.t. average error distance, median error distance and max error distance on all three datasets}. Please note that we leverage MSE Loss as an optimizer target function. MSE loss aims to minimize the whole loss on all data instances (not the median loss or the max loss). Thus we mainly discuss average error distance in this work. The best baseline (i.e., \emph{italic} rows) in the following sections is the baseline with the smallest average error distance on the test sets if there is no special explanation.\\
    \indent
    \textbf{Accuracy}. \textcolor{black}{From Table 1, on training-70\% datasets, GNN-Geo outperforms the best baselines by 28.76\%, 16.61\%, 26.14\%, 21.33\%, 23.16\%, 5.18\%, 25.46\%, 46.73\%, and 9.25\% in NYS (Corr-SLG), HK (MLP-Geo), Shanghai (Corr-SLG), Beijing (SLG), Tokyo (Corr-SLG), Tokyo-IPv6 (MLP-Geo), Berlin-IPv6 (Corr-SLG), and Munich-IPv6 (MLP-Geo) w.r.t. average error distance, respectively. And GNN-Geo also outperforms the best baselines by 30.38\%, 12.32\%, 21.33\%, 25.89\%, 10.79\%, 25.58\%, 36.67\%, and 5.44\% in NYS (Corr-SLG), HK (Corr-SLG), Shanghai (SLG), Beijing (SLG), Tokyo (MLP-Geo), Tokyo-IPv6 (MLP-Geo), Berlin-IPv6 (Corr-SLG), and Munich-IPv6 (MLP-Geo) w.r.t. median error distance, respectively. The best baselines are different in different networks and are shown in brackets. So averagely, in these 8 IPv4/IPv6 networks, GNN-Geo outperforms the best baselines by 22.51\% and 21.05\% w.r.t. average error distance and median error distance, respectively. These results demonstrate that the geolocation accuracy of GNN-Geo is clearly higher than baselines, which validates the value of investigating GNN's potential in fine-grained IP geolocation}.\\
    \indent
    \textbf{Generalization}. \textbf{It is interesting that the best baselines are all different in \textcolor{black}{8 networks}.} \textcolor{black}{For average error distance, the best baseline is Corr-SLG for 4 times, MLP-Geo for 3 times, and SLG for 1 time; for median error distance, the best baseline is Corr-SLG for 3 times, MLP-Geo for 3 times, and SLG for 2 time.} From the experiments, we can see the performances of baselines are not stable in different areas. Their performance varies significantly in different network environments. For SLG, it relies on the proportion of "strong-positive-delay-distance-correlated" landmarks. For Corr-SLG, the proportion of "strong-negative-delay-distance-correlated" landmarks decides whether it can outperform SLG. \textbf{Compared with previous baselines, GNN-Geo shows better generalization capabilities in different network environments}.

\begin{figure*}[htbp]
        \centering
        \subfigure[New York State]{
        \begin{minipage}[t]{0.48\linewidth}
        \centering
        \includegraphics[width=2.6in]{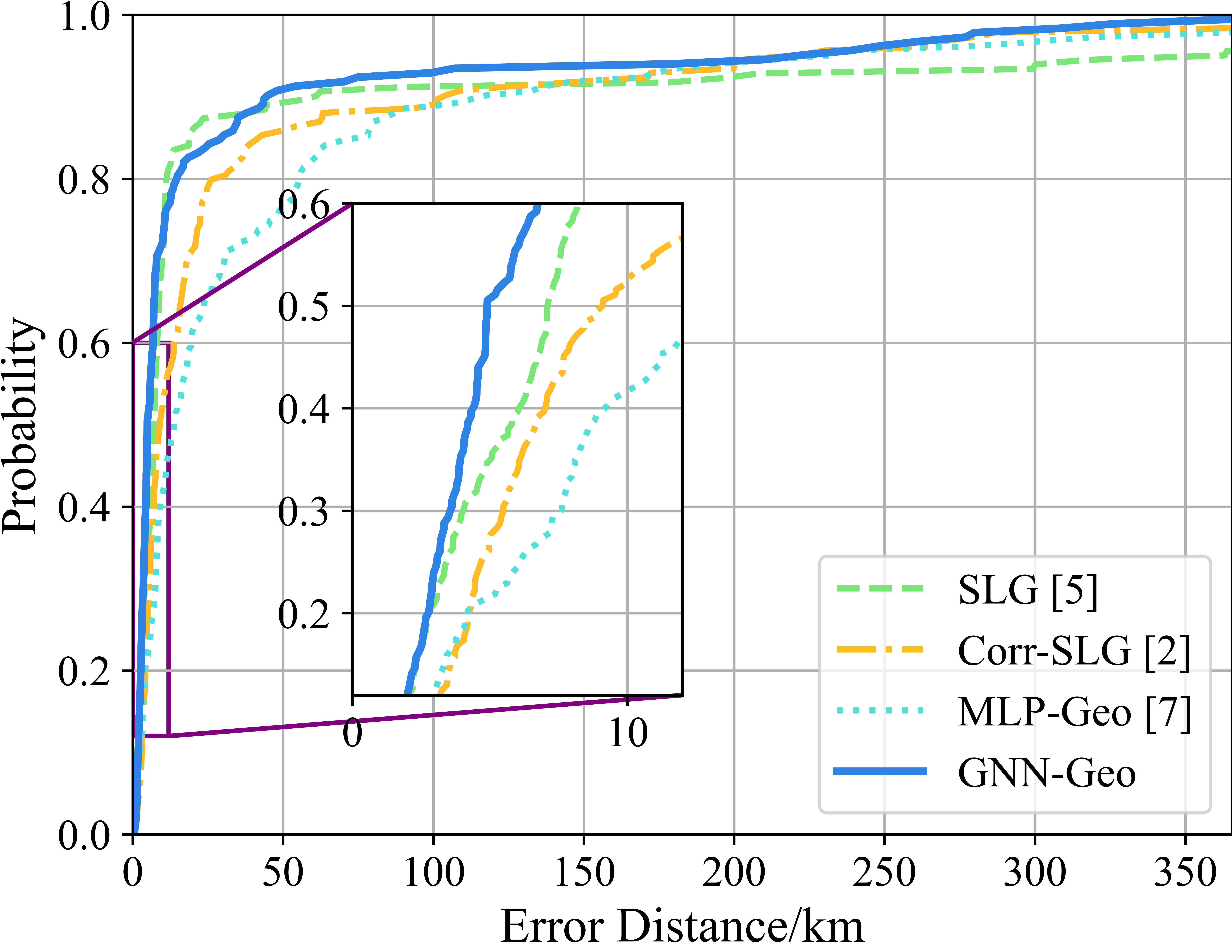}
        %\caption{fig1}
        \end{minipage}%
        \label{fig:ny_cdf}
        }%
        \subfigure[Hong Kong]{
        \begin{minipage}[t]{0.48\linewidth}
        \centering
        \includegraphics[width=2.6in]{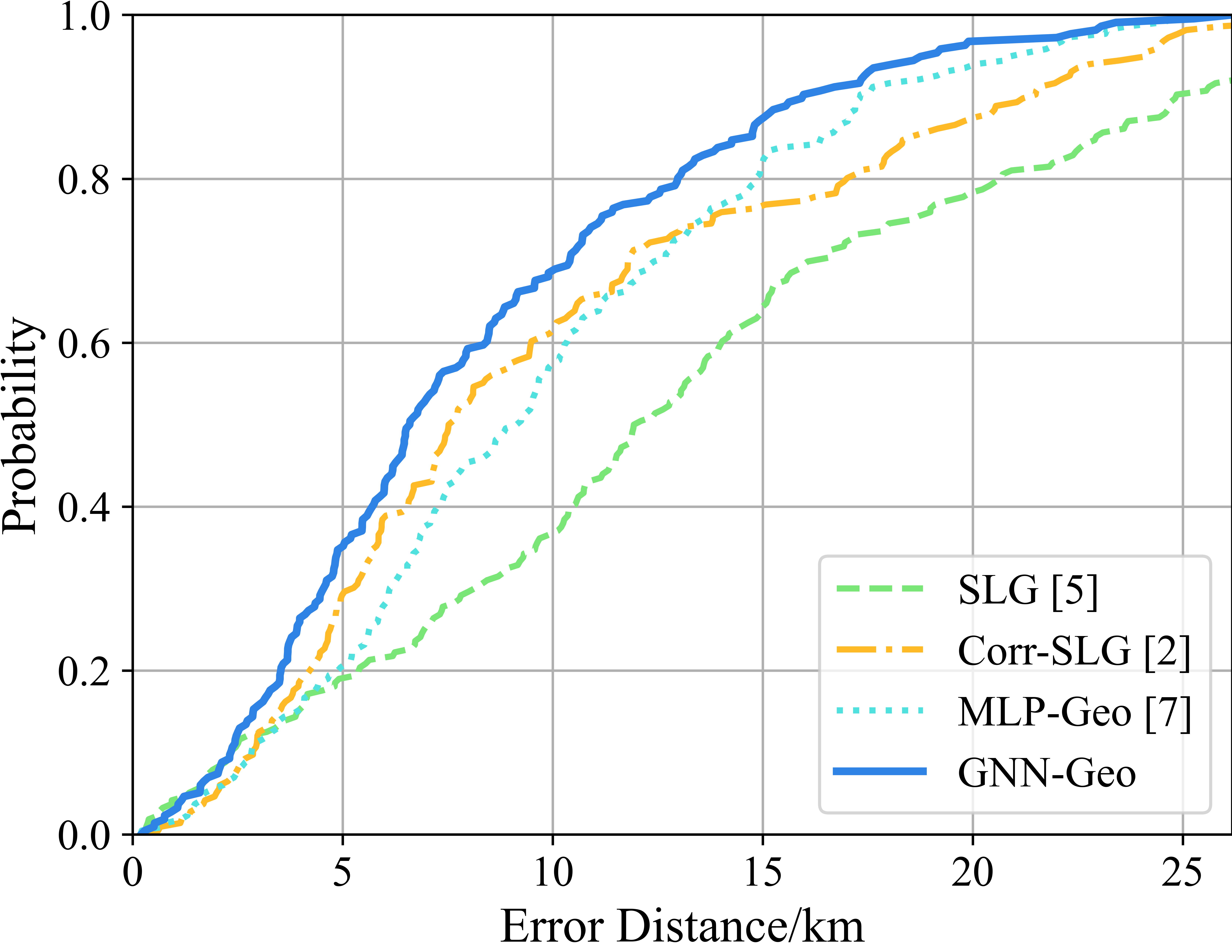}
        %\caption{fig2}
        \end{minipage}%
        \label{fig:hk_cdf}
        }\\
        \subfigure[Shanghai]{
        \begin{minipage}[t]{0.48\linewidth}
        \centering
        \includegraphics[width=2.6in]{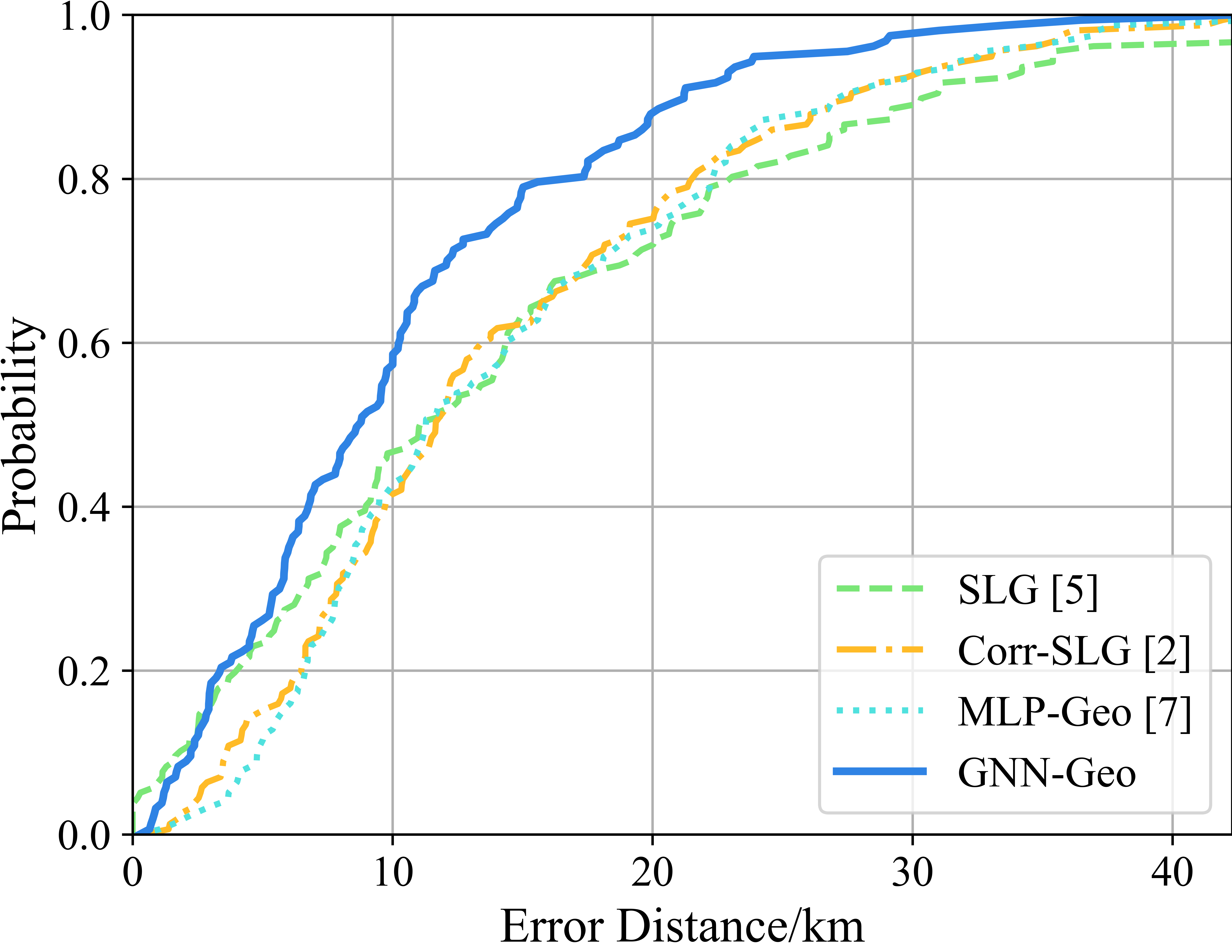}
        %\caption{fig2}
        \end{minipage}%
        \label{fig:sh_cdf}
        }
        \subfigure[Beijing]{
        \begin{minipage}[t]{0.48\linewidth}
        \centering
        \includegraphics[width=2.6in]{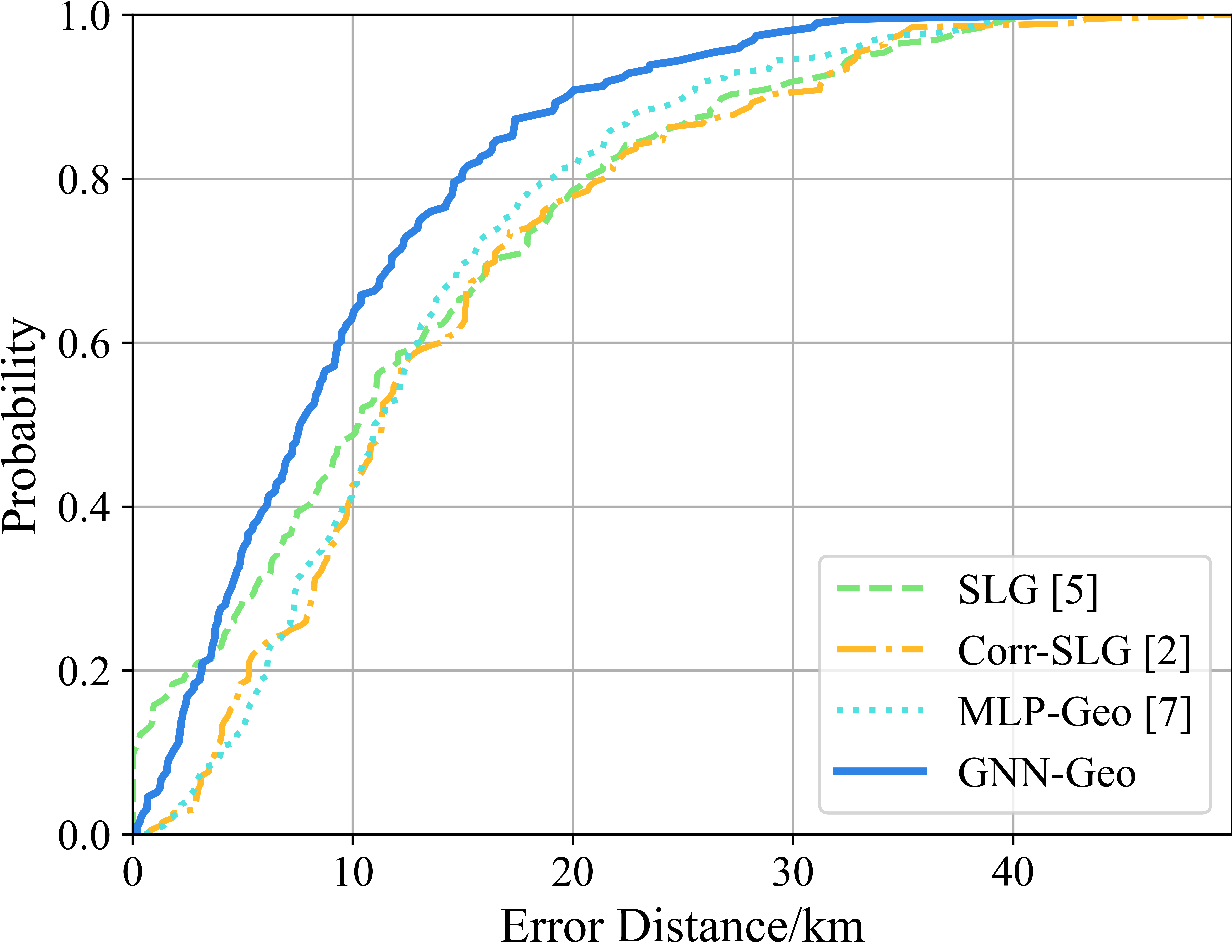}
        %\caption{fig2}
        \end{minipage}
        \label{fig:bj_cdf}
        }\\
        \subfigure[Tokyo]{
        \begin{minipage}[t]{0.48\linewidth}
        \centering
        \includegraphics[width=2.6in]{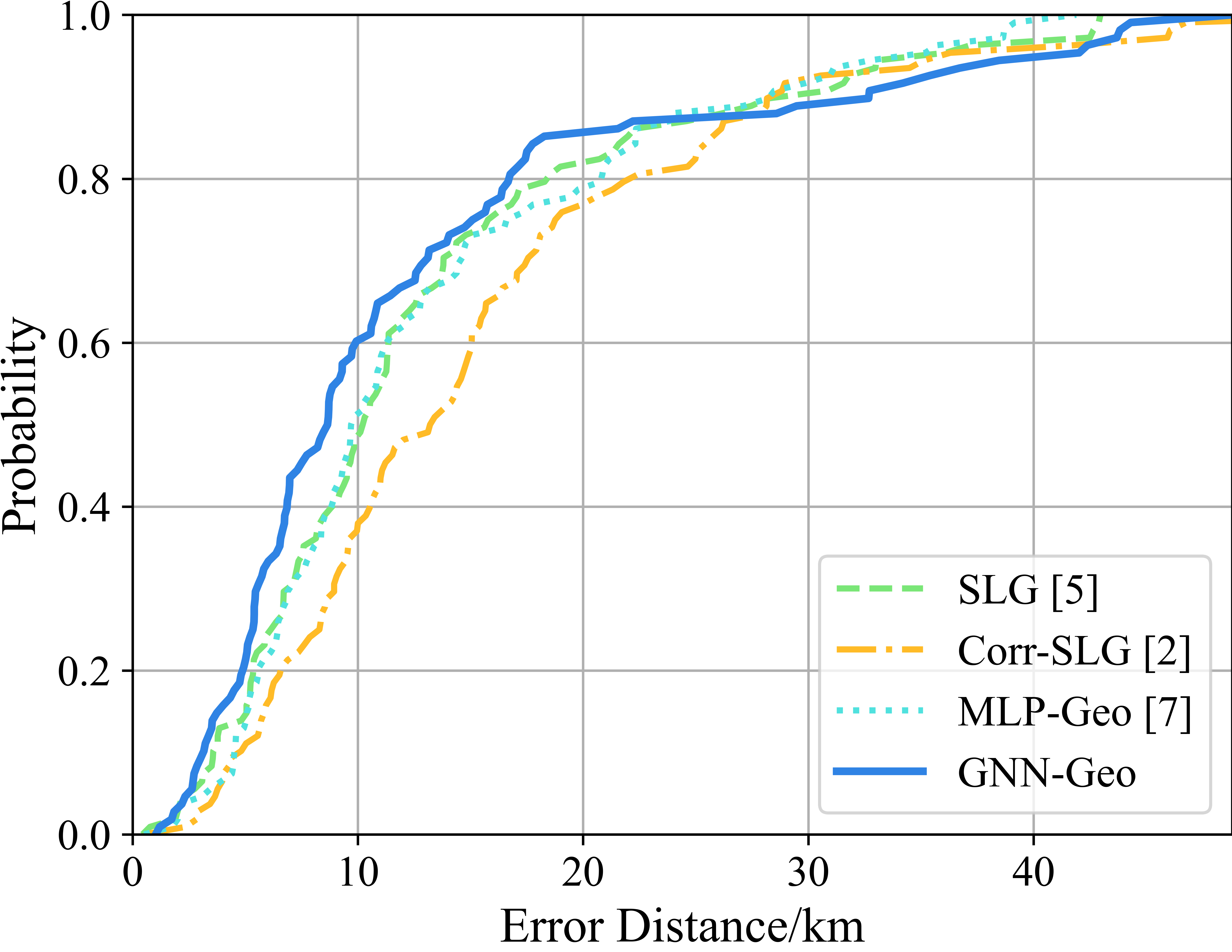}
        %\caption{fig2}
        \end{minipage}%
        \label{fig:dj_cdf}
        }
        \subfigure[Tokyo-IPv6]{
        \begin{minipage}[t]{0.48\linewidth}
        \centering
        \includegraphics[width=2.6in]{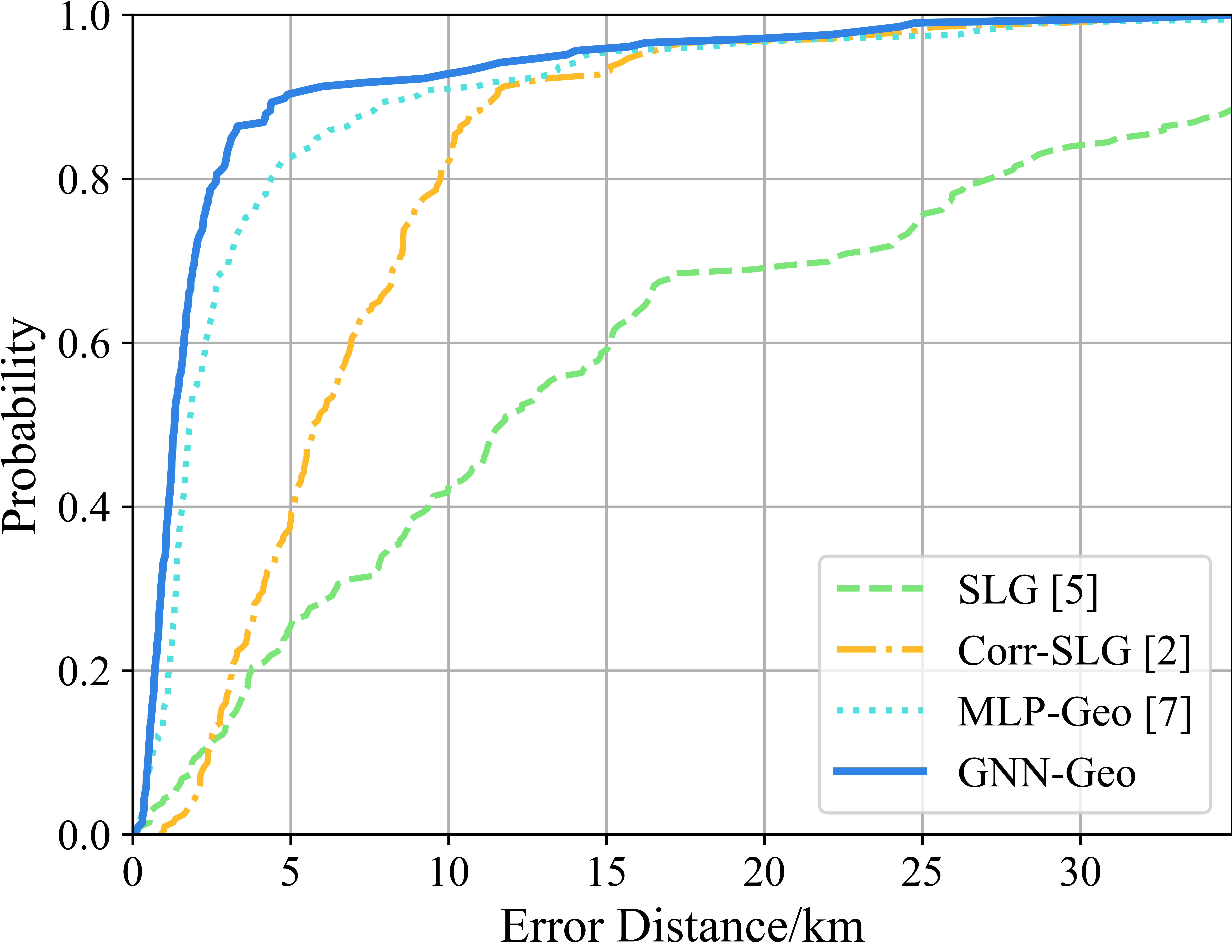}
        %\caption{fig2}
        \end{minipage}
        \label{fig:djv6_cdf2}
        }\\
        \subfigure[Berlin-IPv6]{
        \begin{minipage}[t]{0.48\linewidth}
        \centering
        \includegraphics[width=2.6in]{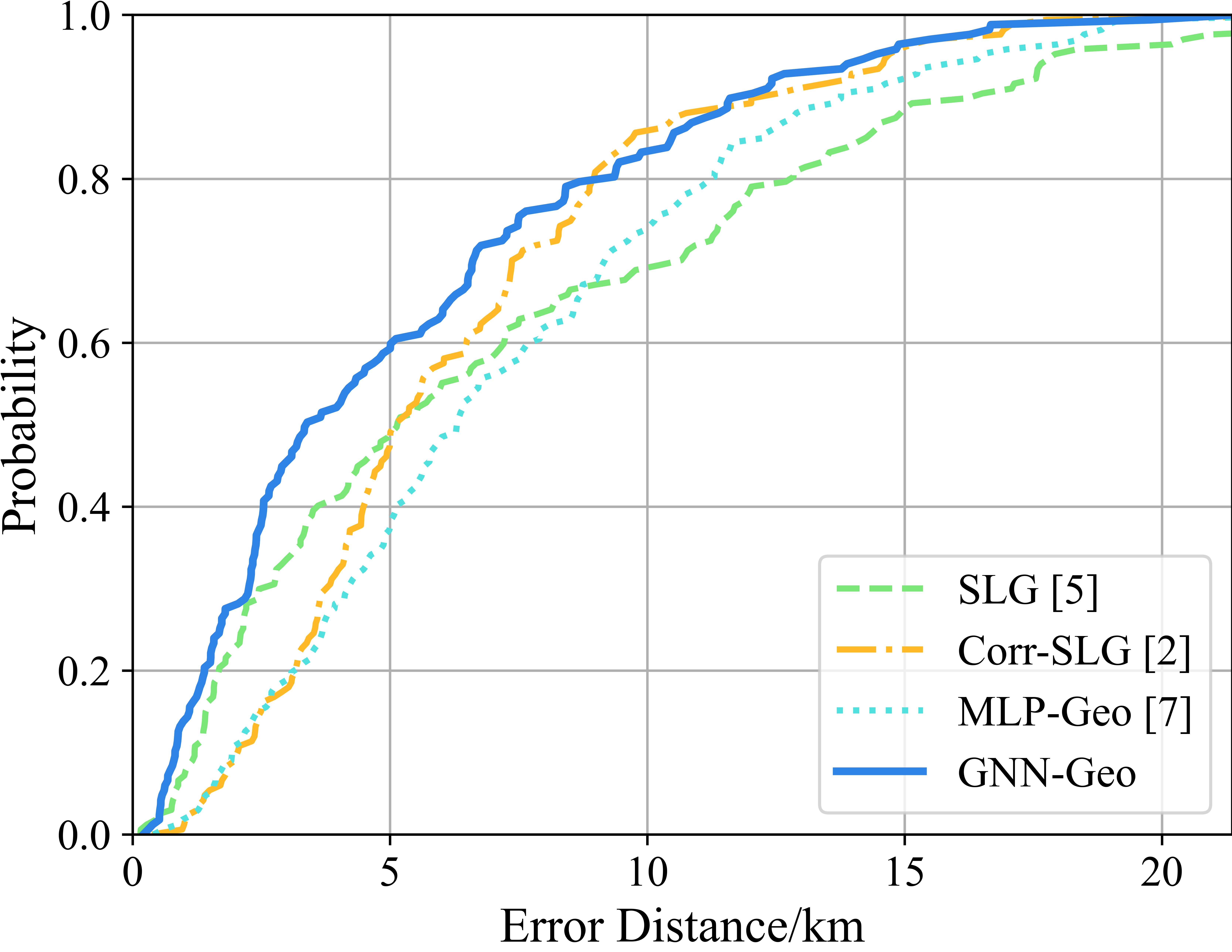}
        %\caption{fig2}
        \end{minipage}%
        \label{fig:bl_cdf}
        }
        \subfigure[Munich-IPv6]{
        \begin{minipage}[t]{0.48\linewidth}
        \centering
        \includegraphics[width=2.6in]{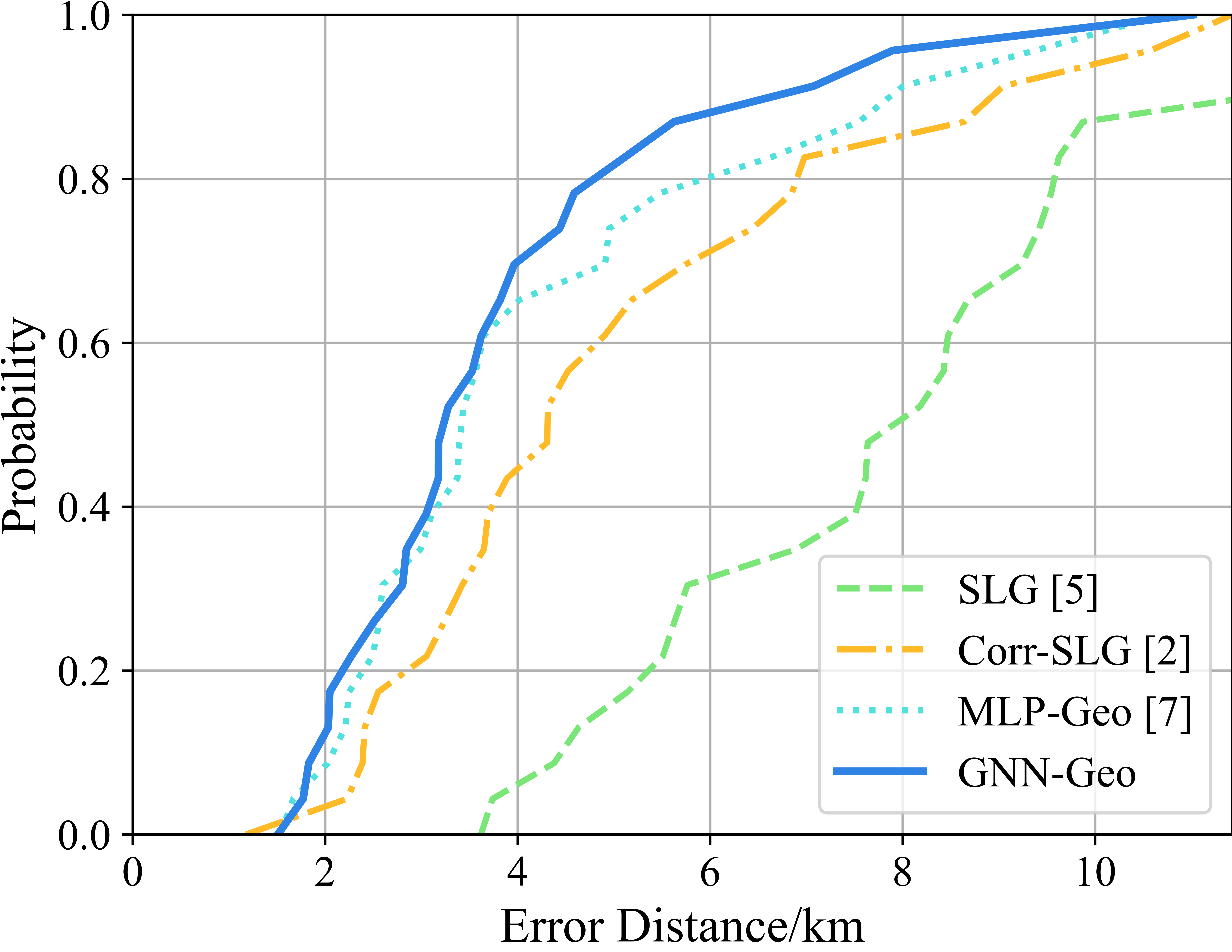}
        %\caption{fig2}
        \end{minipage}
        \label{fig:mnh_cdf}
        }
        \centering
        \caption{\textcolor{black}{Error Distance CDF of 8 Networks (the maximum values of x-axis are GNN-Geo's max error distances)}}
        \label{fig:cdf}
    \end{figure*}
        \begin{figure*}[htbp]
        \centering
        \subfigure[New York State]{
        \begin{minipage}[t]{0.23\linewidth}
        \centering
        \includegraphics[width=1.75in]{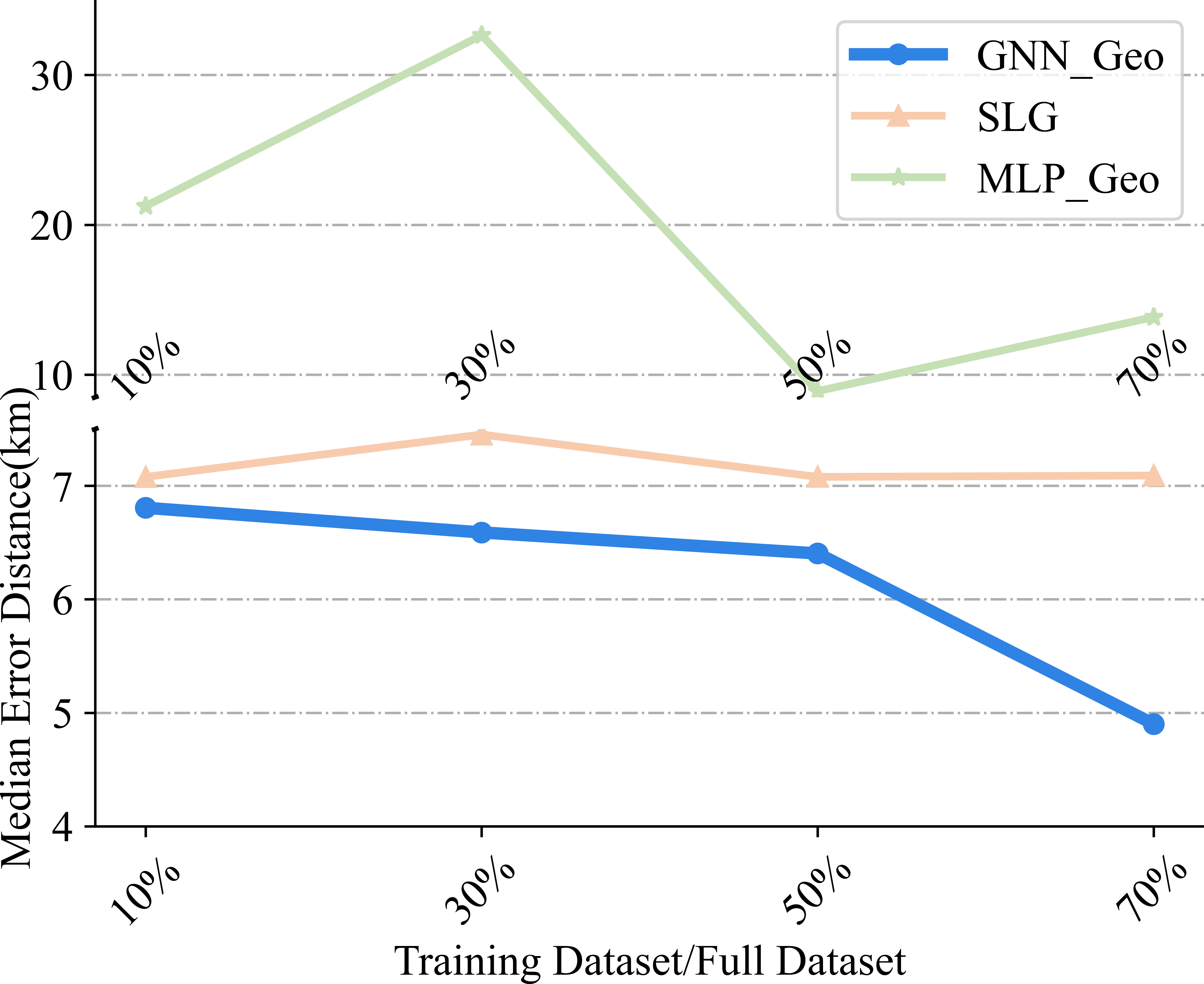}
        %\caption{fig1}
        \end{minipage}%
        \label{fig:ny_m}
        }%
        \subfigure[Hong Kong]{
        \begin{minipage}[t]{0.23\linewidth}
        \centering
        \includegraphics[width=1.75in]{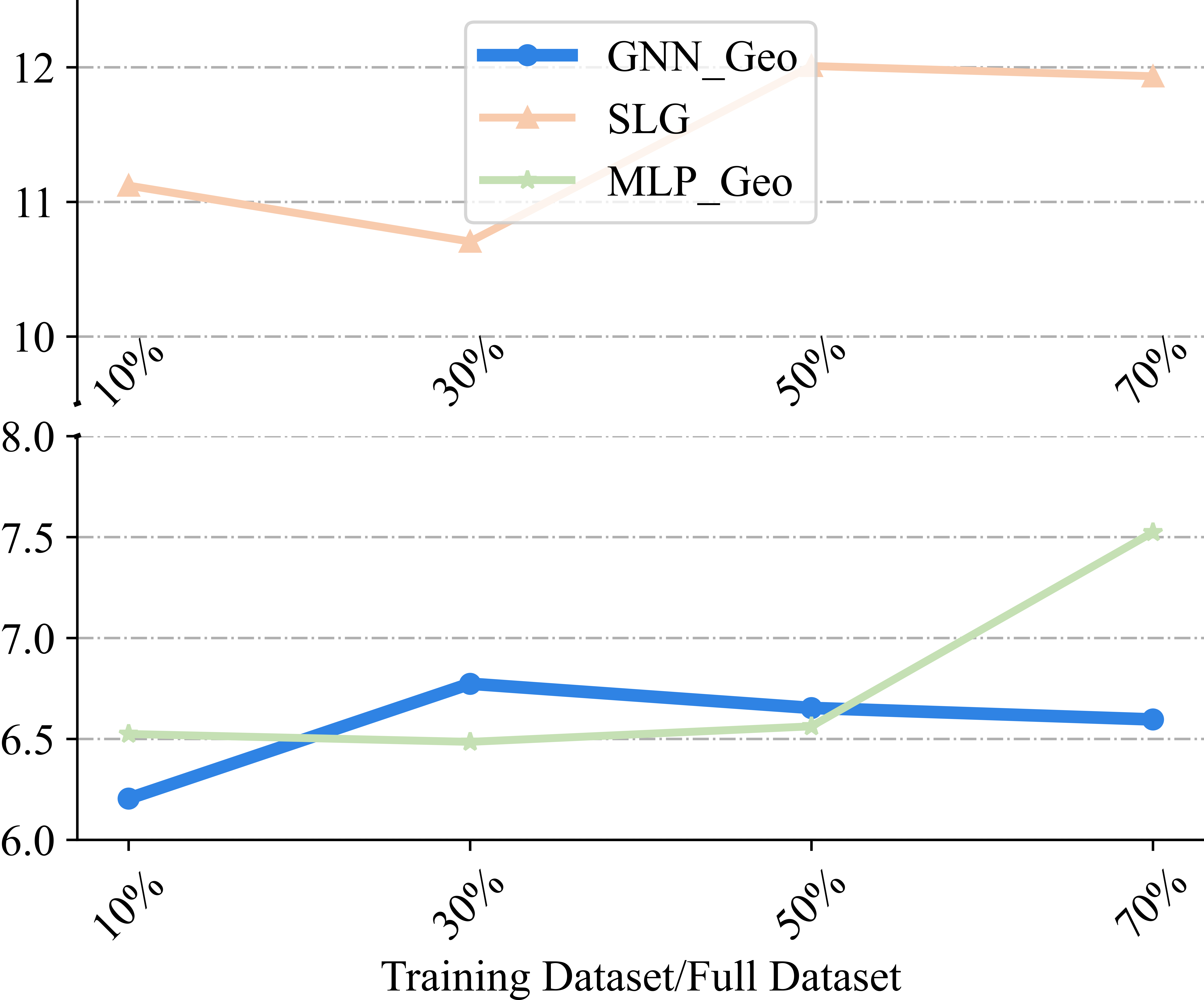}
        %\caption{fig2}
        \end{minipage}%
        \label{fig:hk_m}
        }
        \subfigure[Shanghai]{
        \begin{minipage}[t]{0.23\linewidth}
        \centering
        \includegraphics[width=1.75in]{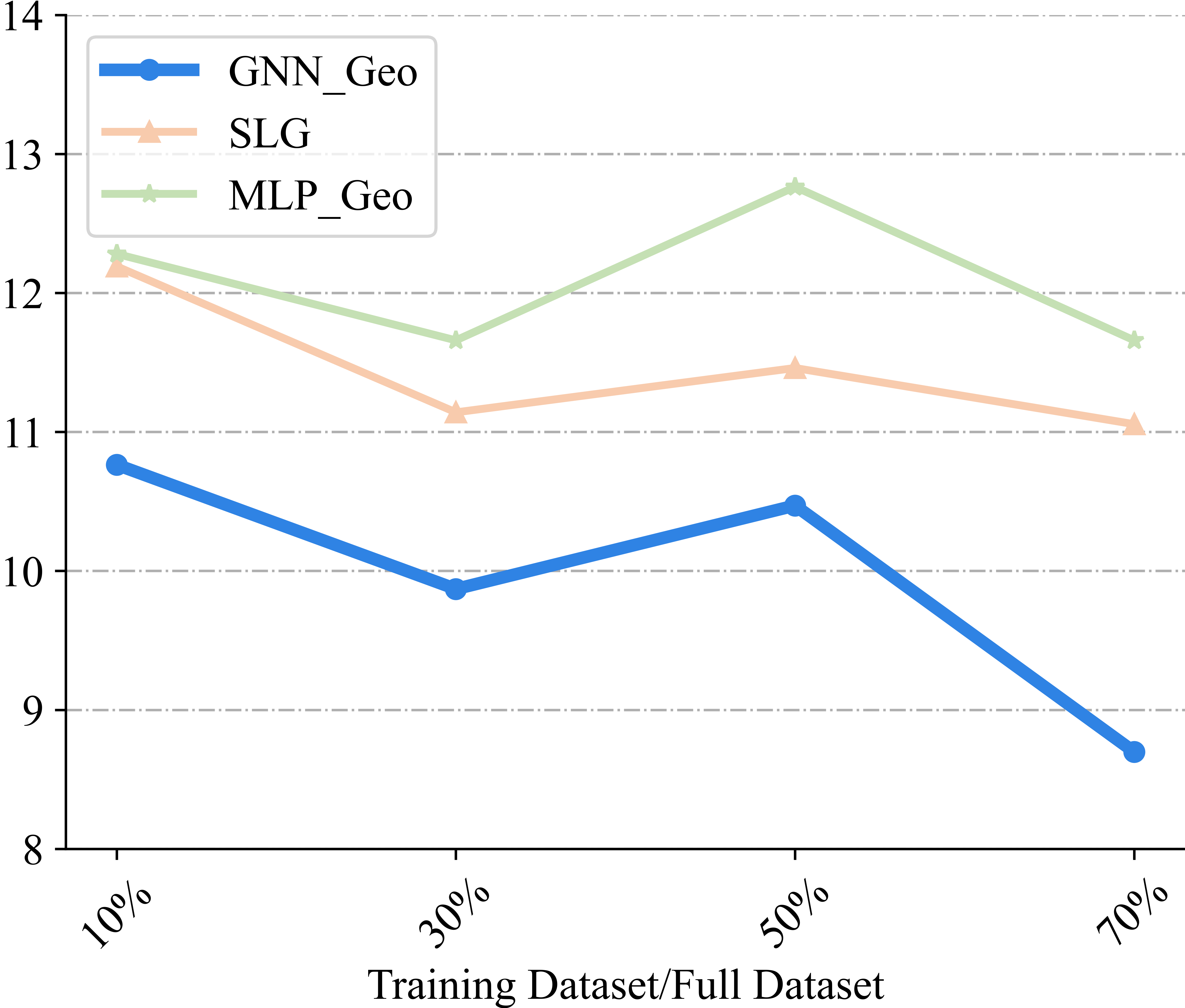}
        %\caption{fig2}
        \end{minipage}%
        \label{fig:sh_m}
        }
        \subfigure[Beijing]{
        \begin{minipage}[t]{0.23\linewidth}
        \centering
        \includegraphics[width=1.75in]{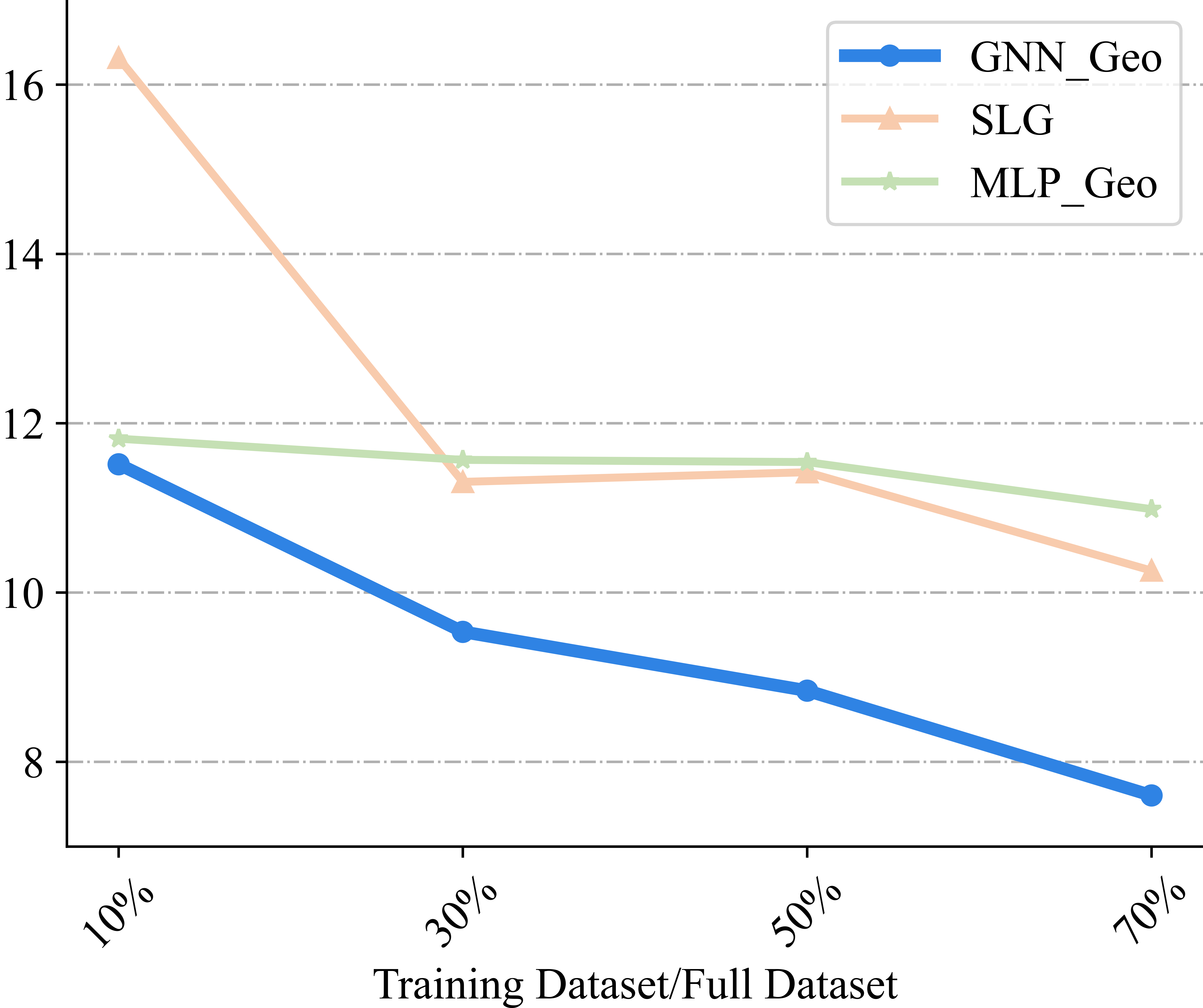}
        %\caption{fig2}
        \end{minipage}
        \label{fig:bj_m}
        }\\
        \subfigure[Tokyo]{
        \begin{minipage}[t]{0.23\linewidth}
        \centering
        \includegraphics[width=1.75in]{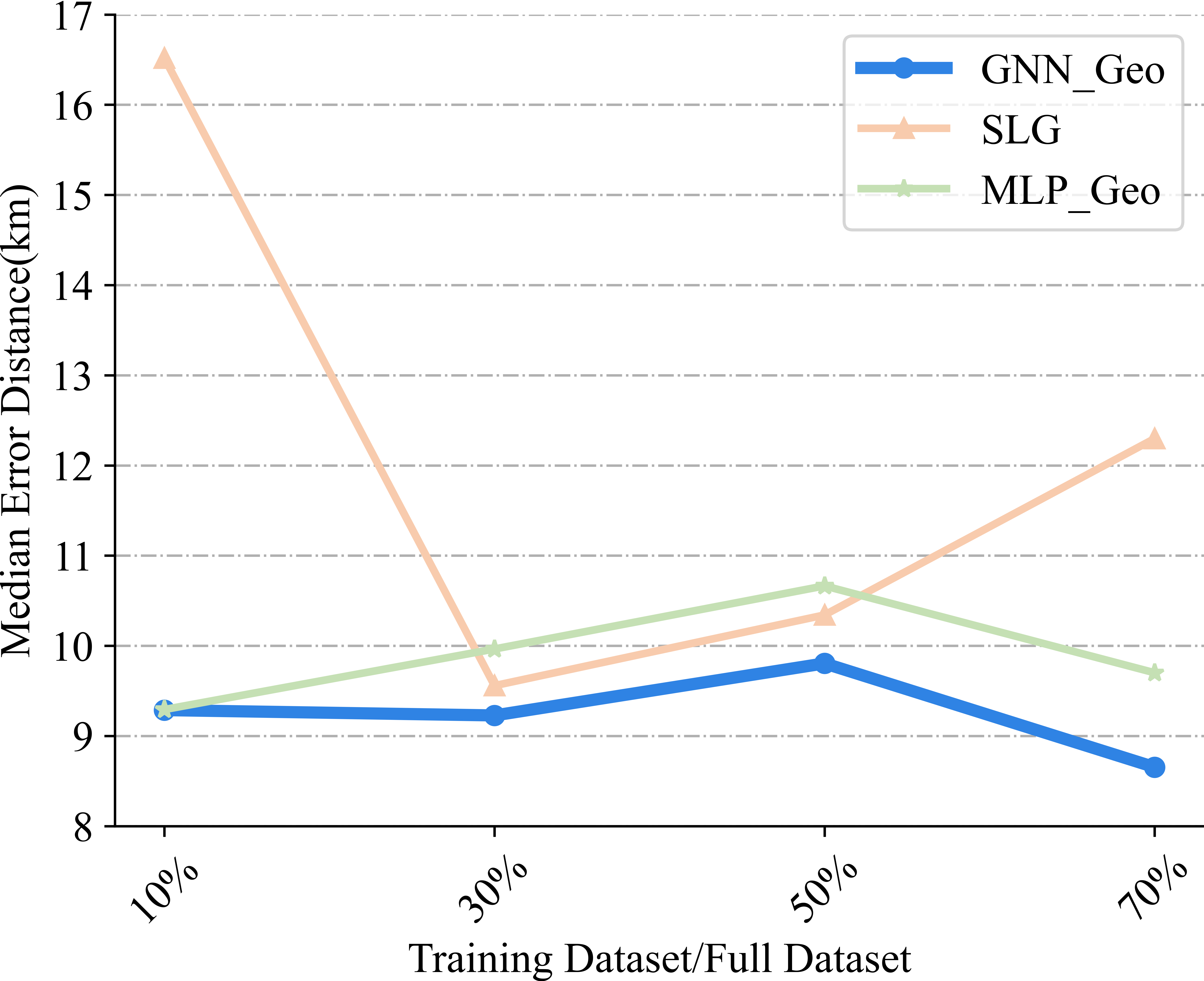}
        %\caption{fig2}
        \end{minipage}%
        \label{fig:dj_m}
        }
        \subfigure[Tokyo (IPv6)]{
        \begin{minipage}[t]{0.23\linewidth}
        \centering
        \includegraphics[width=1.75in]{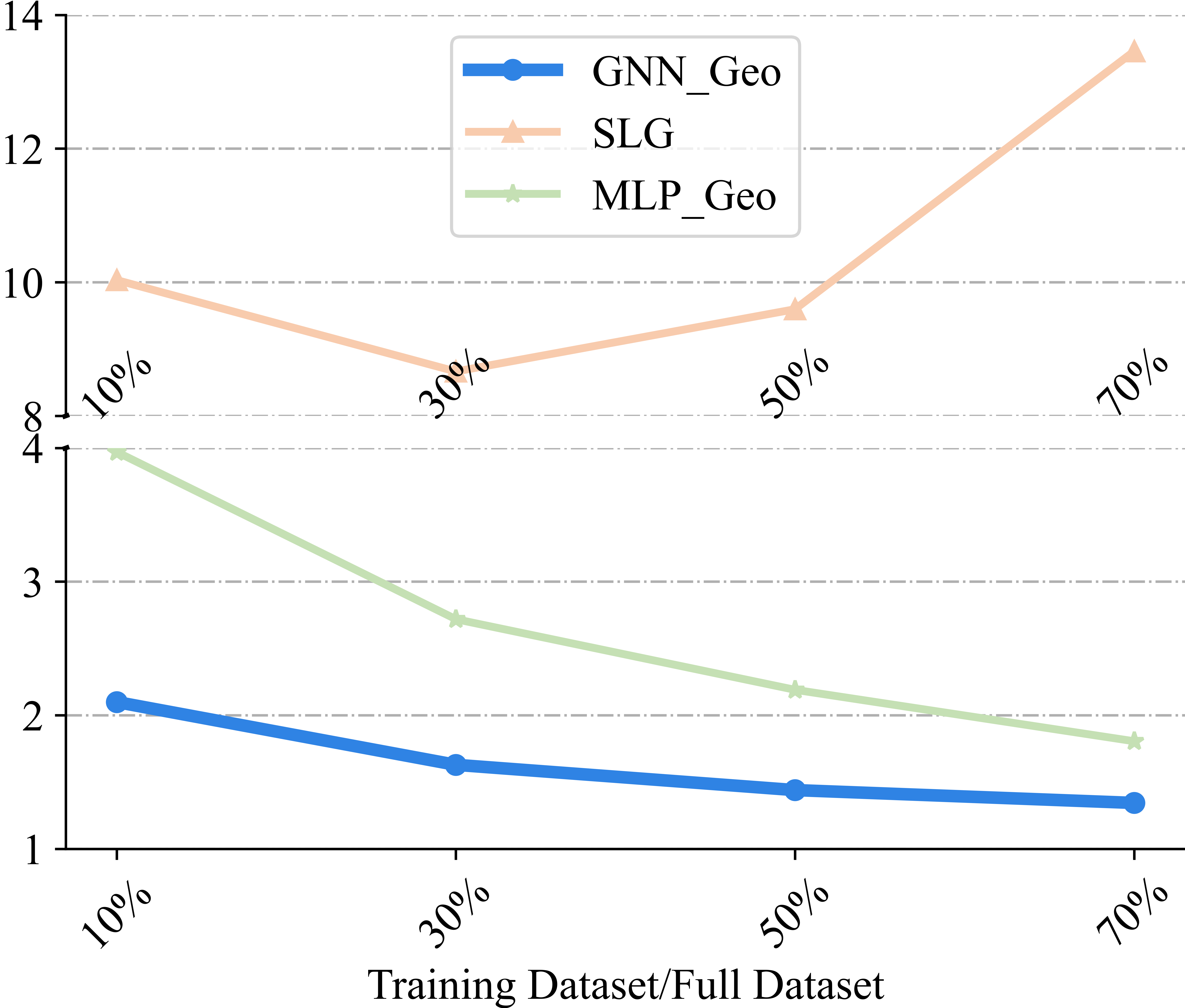}
        %\caption{fig2}
        \end{minipage}
        \label{fig:djv6_m}
        }
        \subfigure[Berlin (IPv6)]{
        \begin{minipage}[t]{0.23\linewidth}
        \centering
        \includegraphics[width=1.75in]{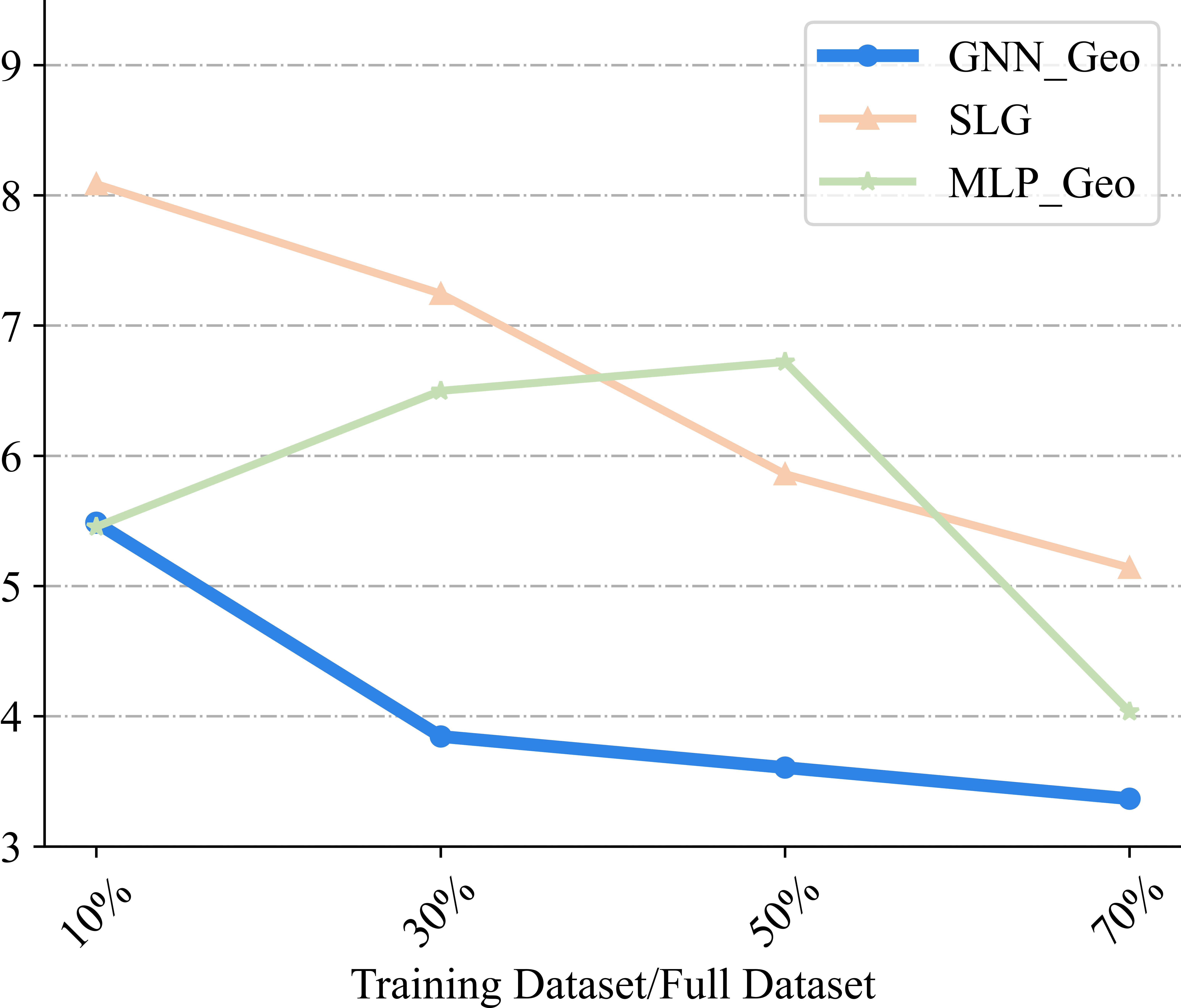}
        %\caption{fig2}
        \end{minipage}%
        \label{fig:bl_m}
        }
        \subfigure[Munich (IPv6)]{
        \begin{minipage}[t]{0.23\linewidth}
        \centering
        \includegraphics[width=1.75in]{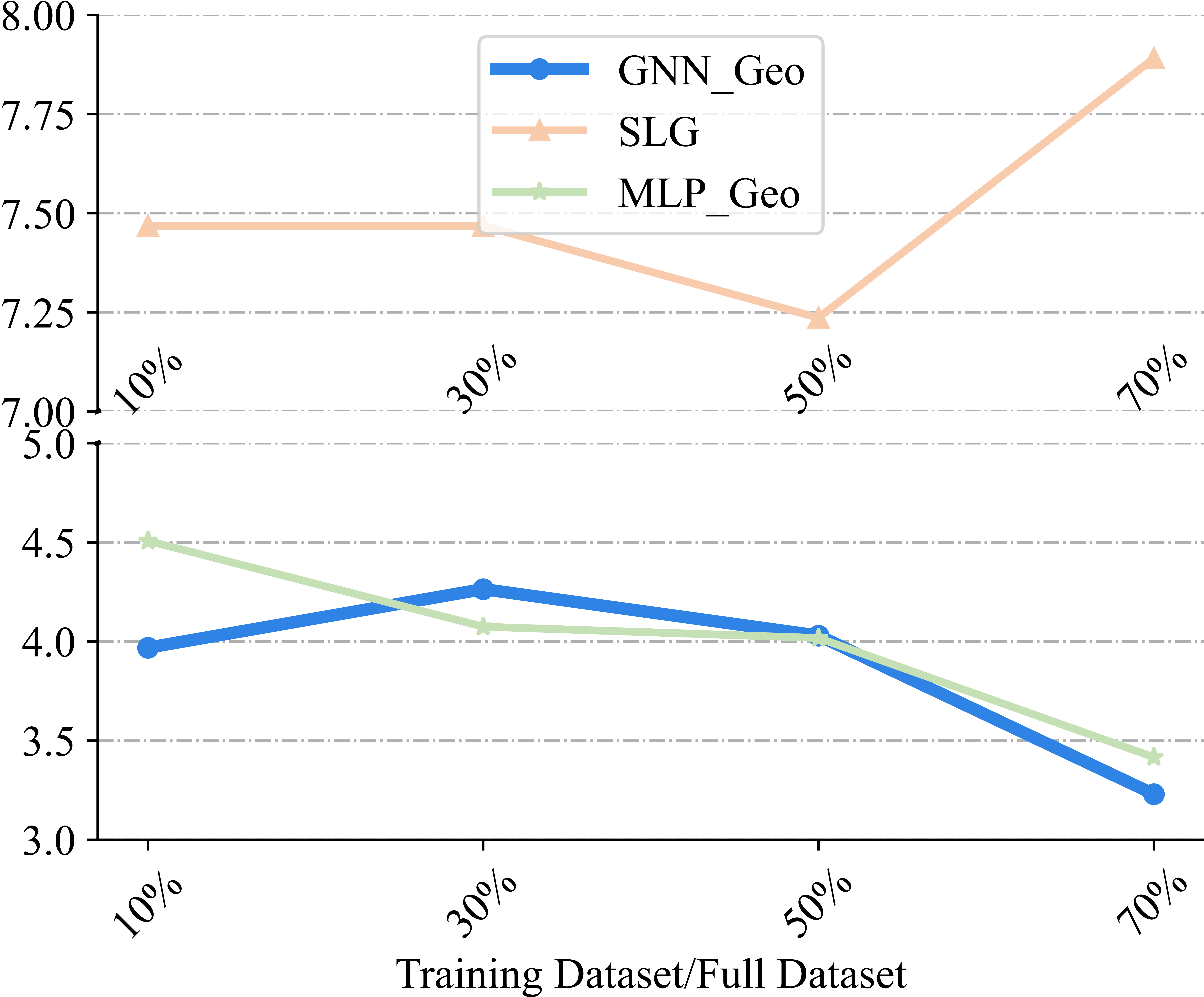}
        %\caption{fig2}
        \end{minipage}
        \label{fig:mnh_m}
        }
        \centering
        \caption{\textcolor{black}{The Relationship between Error Distance and Training dataset/Full dataset}}
        \label{fig:training_ratio}
    \end{figure*}

    \subsection{An Interesting Difference between Learning-based and Rule-based Methods}
    \label{sec:eva_interesting}
    \indent
    Fig. \ref{fig:cdf} shows the cumulative probability of error distances of baselines and GNN-Geo in all \textcolor{black}{8 networks (training-70\% datasets)}. In NYS network, the max error distances are too large, overlapping the smaller cumulative probability with each other. Thus we magnify the picture (0 - 10 km \& 0.1 - 0.6) in \textcolor{black}{Fig. \ref{fig:ny_cdf}}.\\
    \indent
    From \textcolor{black}{Fig. \ref{fig:cdf}}, we can see that: generally, the error distances of GNN-Geo are smaller than baselines in most phases of cumulative probability. \textcolor{black}{This validates GNN-Geo's advantages in most cases.} However, if we check the error distances less than \textcolor{black}{5 km, we can see SLG sometimes could be better than GNN-Geo in the very beginning. For example, in Shanghai and Beijing, SLG is clearly better than GNN-Geo in the smallest error distances (<2km). In fact, some smallest error distances of SLG are even 0 km.} Why SLG could be better than GNN-Geo at the start and become worse in the latter part of cumulative probability?\\%Or in another question, why are the improvements of GNN-Geo compared with SLG usually much larger in average errors than median errors?
    \indent
    The rule of SLG is: "the shortest delay comes from the smallest distance". Thus SLG could get very accurate geolocation results for targets that follow this rule. Some error distances could be close to 0 if the landmarks are at the same location as the targets. This is why SLG performs well in the first part. However, the disadvantage of SLG is that its error distances could be very large for targets that follow the opposite rule -- "the shortest delay comes from the longest distance". Besides, the error distances of SLG are also large for targets which have no clear linear delay-distance relationship at all. This is why the max error distances of SLG are very large in some datasets. \textcolor{black}{If we observe the max error distances when probability reaches 100\%, we can see SLG's max error distances are much larger than other baselines. For example, in Tokyo-IPv6, we could see when GNN-Geo's max error distance is just about SLG's top 85\% max error distance.} So the average error distances of SLG are not so good if there are many of these kinds of targets. The cumulative probability of SLG would quickly become worse in the latter part.\\%: SLG "quit" targets that do not follow its rule
    \indent
    GNN-Geo does not follow a specific linear delay-distance rule. Its "measurement data - location" mapping function is learned based on a supervised-learning method. The mapping function is optimized by minimizing the MSE loss of the whole training set. In other words, GNN-Geo considers the whole performance or average performance of all targets. Its mapping function tries to be suitable to as many targets as possible, instead of only focusing on a part of targets. In a dataset, if different targets follow different delay-distance or topology-distance rules, the mapping function of GNN-Geo may be less accurate for some targets than SLG. This is because the mapping function needs to get better performance for other targets which do not follow linear rules. Therefore GNN-Geo can outperform SLG in the latter part of cumulative probability. Actually, we can notice that MLP-Geo follows a similar trend with GNN-Geo. In \textcolor{black}{Fig. \ref{fig:sh_cdf}}, MLP-Geo is worse than SLG before 3 km and quickly becomes better afterward. This is because MLP-Geo is also a learning-based method and is optimized by MSE loss.

    \subsection{\textcolor{black}{How training ratios affect GNN-Geo's Performance}}
    \textcolor{black}{Fig. \ref{fig:training_ratio} shows how different methods' performances are affected by changing training-validation-testing ratios in all datasets. Here, we select a typical method in rule-based learning (SLG) and learning-based learning methods (MLP-Geo) to compare with GNN-Geo. Parameters are re-tuned for learning-based methods when changing ratios. We notice that, though average error distances generally follow similar trends with median error distances, average error distances are easier to be affected by max error distance. Since the training ratio become smaller, some methods may have too large average/max error distances, which make other results in the same figures hard to distinguish. So, we mainly use median error distance to show the trends in Fig. \ref{fig:training_ratio}.}

    \begin{table*}[htbp]
        \label{tab:decoder}
        \centering
        \renewcommand\arraystretch{1.2}
        \caption{Performance (kilometers) Comparison of Different Decoder Types}
        \begin{threeparttable}
        \begin{tabular}{cc|cc|cc|cc}
        \hline
        \multicolumn{2}{c|}{\multirow{2}{*}{Decoder Type}}        & \multicolumn{2}{c|}{New York \textcolor{black}{State}}   & \multicolumn{2}{c|}{Hong Kong} & \multicolumn{2}{c}{Shanghai}   \\
        \multicolumn{2}{c|}{}                                     & Average         & Epochs        & Average         & Epochs       & Average        & Epochs        \\ \hline
        \multirow{2}{*}{\textcolor{black}{Pure MLP-based Decoders}}   & Vanilla MLP                  & 3406.247        & 512           & 2690.566        & 10000         & 3266.079       & 10000          \\
                                    & +BatchNorm                  & 6010.929        & 26            & 2812.680        & 385          & 1087.136       & 20            \\ \hline
        \multirow{4}{*}{\textcolor{black}{MLP-based Decoders with Rules}} & Vanilla MLP                  & 69.663          & 6643          & 8.127           & 9828         & 12.504         & 3589          \\
                                    & +BatchNorm                  & 29.861          & 2073          & 7.735           & 4690         & {\ul 9.723}    & 1808          \\
                                    & +Sigmoid                    & {\ul 26.445}    & 1484          & {\ul 7.681}     & 2265         & 9.843          & 1096          \\
                                    & \textbf{+BatchNorm+Sigmoid} & \textbf{25.165} & \textbf{3534} & \textbf{7.603}  & \textbf{760} & \textbf{9.673} & \textbf{1439} \\ \hline
        \end{tabular}

        \begin{tablenotes}
            \item[1] \textbf{Bold} indicate the average error distance of GNN-Geo, {\ul underline} rows indicate the validation average error distances among other decoder types. Note all these methods only differ in decoder types.
            \item[2] "Epochs" indicates the training epoch number of the corresponding results.
            \item[3] These results are the best average distance errors on validation sets.
        \end{tablenotes}
        \end{threeparttable}
    \end{table*}

    \textcolor{black}{From Fig. \ref{fig:training_ratio}, we can see that the performances of all methods generally become better when training ratio increases from 10\% to 70\%, validating that more landmarks in the same network usually lead to better geolocation accuracy for measurement-based methods. In 18 out of 24 cases (75\%), GNN-Geo still clearly outperforms baselines with training ratio less than 70\%. This shows that we still can use GNN-Geo even with relatively less landmarks. However, if the training ratio is too less (training-ratio is 10\%), only in 50\% cases, GNN-Geo is clearly better than other baselines. This shows that, compared with other methods, GNN's performances may be more easily affected with extremely smaller training datasets. This is because: (1) compared with SLG, as a learning-based method, GNN-Geo could be overfit on extremely sparse datasets; (2) compared with MLP-Geo, GNN-Geo has more parameters to train, so it is generally easier to overfit. So sometimes, if the training dataset is too small, and the signals (e.g., network topology) extracted by GNN-Geo are not so important to IP geolocation, then GNN-Geo's advantages could be decreased.}

    \subsection{Geolocation Performance Comparison over different Decoders}%Overall
    \label{sec:Performance_Comparison_over_GCN-Geo-baselines}
    As explained in section \ref{sec:model_decoder}, GNN-based geolocation methods with the original decoders may be not suitable for IP geolocation. Here we compare the proposed decoder with several baseline decoders. All these decoders share the same preprocessor, encoder and MP layers. \textcolor{black}{Pure MLP-based} decoders do not limit the output data range. \textcolor{black}{MLP-based decoders improved by rules} scale the latitude and longitude range of the training set (+11 km) to [0,1]. For \textbf{Vanilla MLP}, the decoder consists of a dense layer with Relu and an output layer. For \textbf{+BatchNorm}, the decoder consists of a dense layer with Relu, a batch normalization layer and an output layer. For \textbf{+Sigmoid}, the decoder consists of a dense layer with Relu and an output layer with Sigmoid. For \textbf{+BatchNorm+Sigmoid}, the decoder consists of a dense layer with Relu, a batch normalization layer and an output layer with Sigmoid. \textbf{+BatchNorm+Sigmoid} is the proposed rule-based decoder used in GNN-Geo.\\
    \indent
    The best geolocation performance w.r.t average error distance of all decoders on validation sets is shown in Table 2. \textcolor{black}{Here we only show the results in first 3 networks (NYS, HK, Shanghai) with training-70\% ratio as an example}. For each dataset, the \textbf{bold} indicates the best average error distance for all decoders while the \emph{italic} row indicates the best average error distance for all baseline decoders. "Epochs" indicate the training epoch number of the corresponding results. It shows how many epochs a decoder needs to converge. The training epochs are at most 10000 and early stopped if results are not improved in 1000 epochs.\\%Among 5 times of train-validation-testing split dataset, we only show the performances and epoch numbers of one splitting due to page limits. The other 4 split dataset follow similar trends.
    \indent
    From Table 2, we can see the best average distance errors of two pure \textcolor{black}{MLP-based decoders} are more than 1,000 km even after 10,000 training epochs. This is because most of the output data belong to $\mathbb{R}$ and are far away from the real locations of the targets. \textbf{It indicates that GNN is not a "silver bullet" to the IP geolocation problem}. From Table \ref{tab:mainresults}, we can the average distances of GNN-Geo are between 8 km and 28 km, clearly outperforming all baselines. However, simply changing its decoder could lead to unacceptable performance. \textbf{The original GNN cannot be directly used for IP geolocation}. We still need to carefully design the structure of the whole GNN-Geo framework to gain better geolocation performance.\\
    \indent
    The performances of \textcolor{black}{decoders improved with rules} are much better. We can find some trends: 1) \textbf{+Sigmoid} and \textbf{+BatchNorm} both can clearly increase the performance and significantly reduce the epoch numbers for convergence; 2) \textbf{+Sigmoid} is more accurate than \textbf{+BatchNorm} except Shanghai dataset; 3) \textbf{+Sigmoid} converges faster than \textbf{+BatchNorm}; 4) \textbf{+BatchNorm+Sigmoid} is better than only using \textbf{+BatchNorm} or \textbf{+Sigmoid}. We explain these trends as follows. Sigmoid limits the range of output into $[0,1]$, which eases the burden for the optimizer. All the output data must be in the area of the training datasets, which is a coarse-grained location estimation of the targets. However, only using Sigmoid could meet the saturating problem. Batch normalization can help to relieve this problem. Batch normalization itself can also facilitate neural network training \cite{ioffe2015batch,santurkar2018does}. Besides, it is noticed that rule-based Vanilla MLP is much better than the original Vanilla MLP, with its output still belonging to $\mathbb{R}$. This is because our initializers sometimes would make the decoder's estimated data in the first several epochs close to 0. If this happens, its performance could be much better than the original MLP. These findings show that we can combine the advantages of rule-based and learning-based methods together to gain better performance.\\
    \indent
    These findings and the advantages of SLG shown in Section \ref{sec:eva_interesting} indicate that we should consider combining the advantages of rule-based and learning-based methods together to gain better performance.
    %%\ref{tab:decoder}

    \begin{figure*}[htbp]
        \centering
        \subfigure[Average Error(km)]{
        \begin{minipage}[t]{0.31\linewidth}
        \centering
        \includegraphics[width=2.1in]{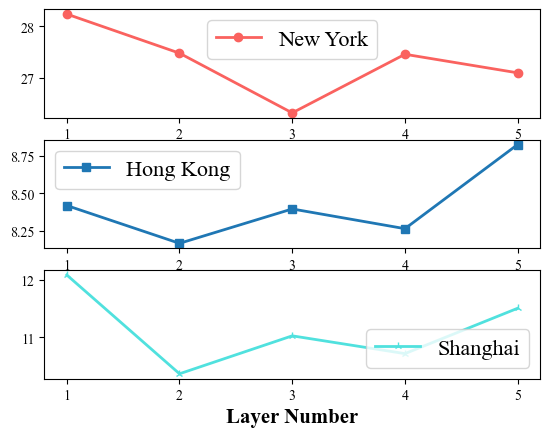}
        %\caption{fig1}
        \end{minipage}%
        \label{fig:para_layer_average}
        }%
        \subfigure[Median Error(km)]{
        \begin{minipage}[t]{0.31\linewidth}
        \centering
        \includegraphics[width=2.1in]{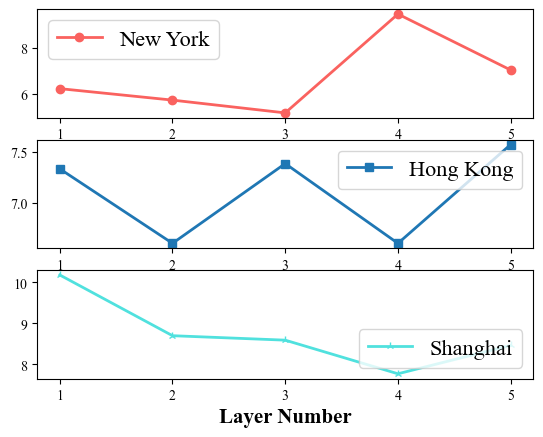}
        %\caption{fig2}
        \end{minipage}%
        \label{fig:para_layer_median}
        }
        \subfigure[Max Error(km)]{
        \begin{minipage}[t]{0.31\linewidth}
        \centering
        \includegraphics[width=2.1in]{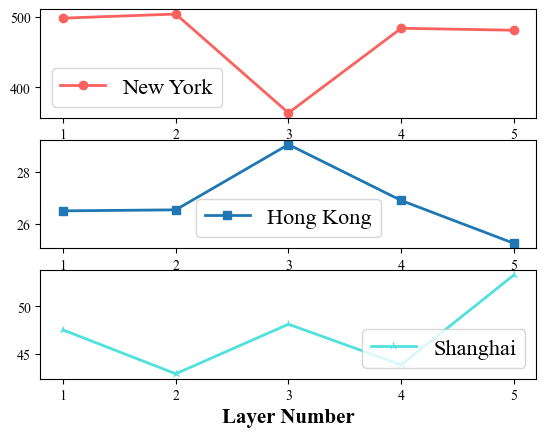}
        %\caption{fig2}
        \end{minipage}%
        \label{fig:fig:para_layer_max}
        }\\%%%%%%%%%%%%%%%%%%%%%%%%%%%%%%%%%%%%%%%%%%%%%%%%%%%%%

        \subfigure[Average Error(km)]{
        \begin{minipage}[t]{0.31\linewidth}
        \centering
        \includegraphics[width=2.1in]{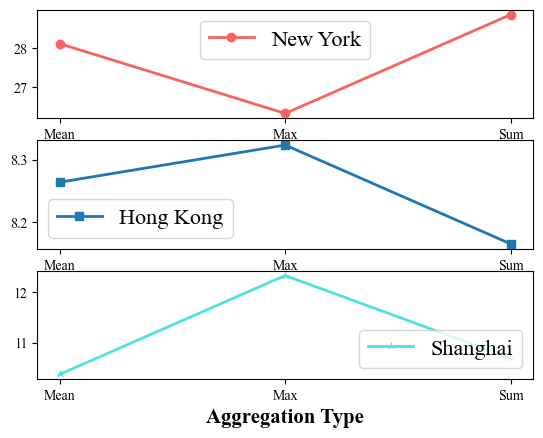}
        %\caption{fig1}
        \end{minipage}%
        \label{fig:para_agg_average}
        }%
        \subfigure[Median Error(km)]{
        \begin{minipage}[t]{0.31\linewidth}
        \centering
        \includegraphics[width=2.1in]{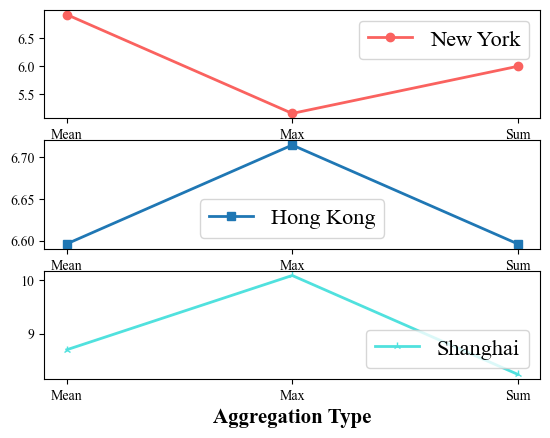}
        %\caption{fig2}
        \end{minipage}%
        \label{fig:para_agg_median}
        }
        \subfigure[Max Error(km)]{
        \begin{minipage}[t]{0.31\linewidth}
        \centering
        \includegraphics[width=2.1in]{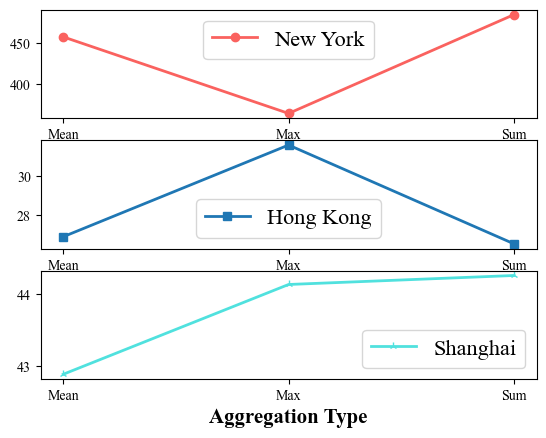}
        %\caption{fig2}
        \end{minipage}%
        \label{fig:para_agg_max}
        }\\%%%%%%%%%%%%%%%%%%%%%%%%%%%%%%%%%%%%%%%%%%%%%%%%%%%%%

        \subfigure[Average Error(km)]{
        \begin{minipage}[t]{0.31\linewidth}
        \centering
        \includegraphics[width=2.1in]{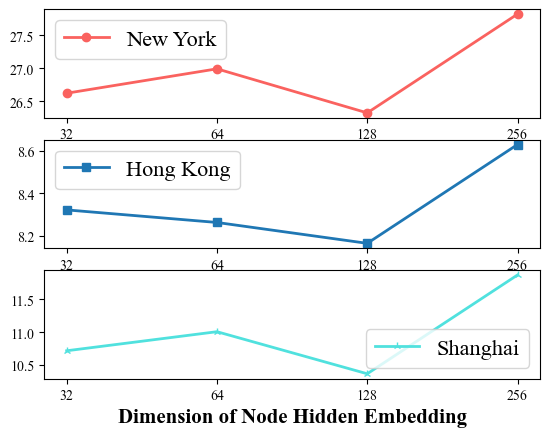}
        %\caption{fig1}
        \end{minipage}%
        \label{fig:para_nodeem_average}
        }%
        \subfigure[Median Error(km)]{
        \begin{minipage}[t]{0.31\linewidth}
        \centering
        \includegraphics[width=2.1in]{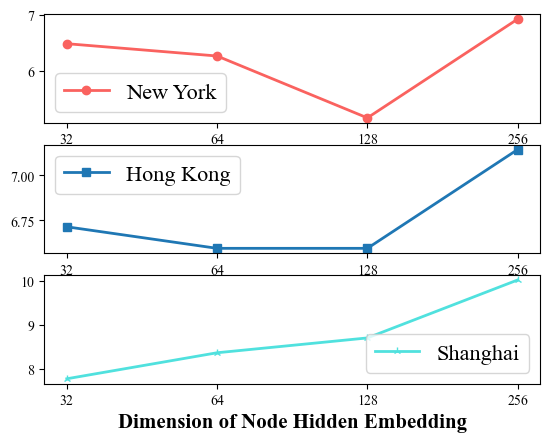}
        %\caption{fig2}
        \end{minipage}%
        \label{fig:para_nodeem_median}
        }
        \subfigure[Max Error(km)]{
        \begin{minipage}[t]{0.31\linewidth}
        \centering
        \includegraphics[width=2.1in]{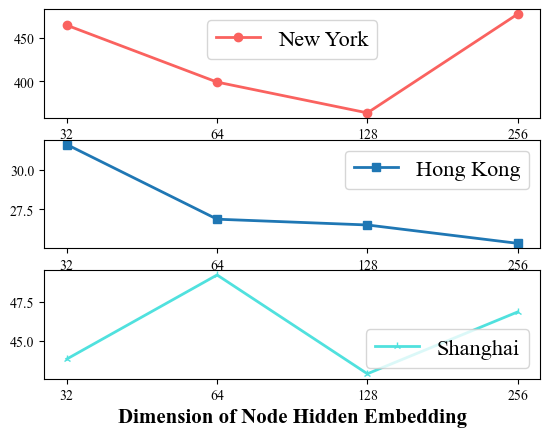}
        %\caption{fig2}
        \end{minipage}%
        \label{fig:para_nodeem_max}
        }\\%%%%%%%%%%%%%%%%%%%%%%%%%%%%%%%%%%%%%%%%%%%%%%%%%%%%%

        \subfigure[Average Error(km)]{
        \begin{minipage}[t]{0.31\linewidth}
        \centering
        \includegraphics[width=2.1in]{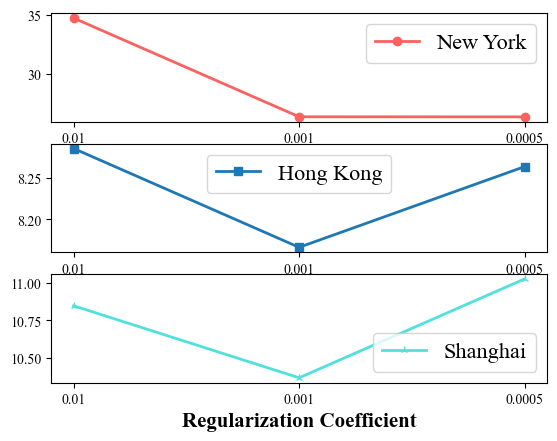}
        %\caption{fig1}
        \end{minipage}%
        \label{fig:para_reg_average}
        }%
        \subfigure[Median Error(km)]{
        \begin{minipage}[t]{0.31\linewidth}
        \centering
        \includegraphics[width=2.1in]{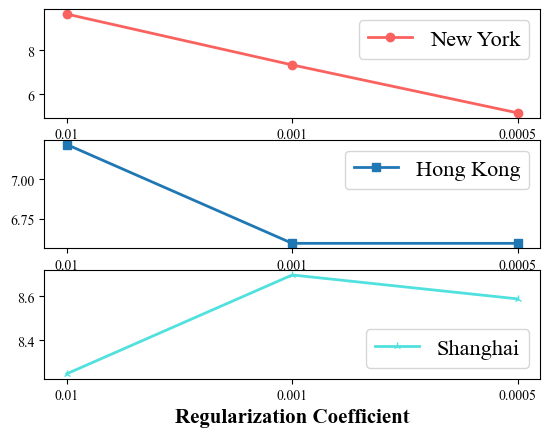}
        %\caption{fig2}
        \end{minipage}%
        \label{fig:para_reg_median}
        }
        \subfigure[Max Error(km)]{
        \begin{minipage}[t]{0.31\linewidth}
        \centering
        \includegraphics[width=2.1in]{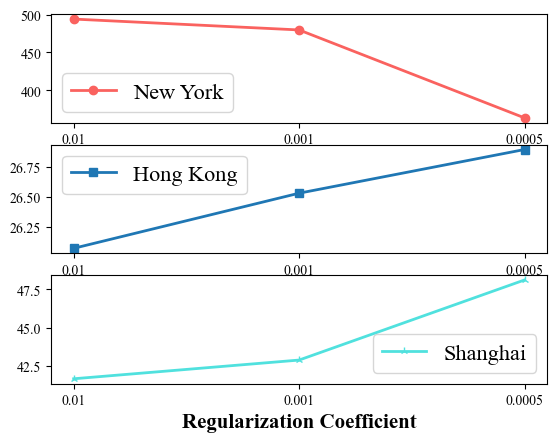}
        %\caption{fig2}
        \end{minipage}%
        \label{fig:para_reg_max}
        }\\%%%%%%%%%%%%%%%%%%%%%%%%%%%%%%%%%%%%%%%%%%%%%%%%%%%%%
        \centering
        \caption{Parameter Study of GNN-Geo in New York State, Hong Kong, Shanghai}
        \label{fig:para}
    \end{figure*}

    \subsection{Further Investigation on GNN-Geo Properties}
    In this section, we investigate the impact of different learning configurations on the performance of GNN-Geo. We start by exploring the influence of the depth of the MP layers. We then study how different message aggregation methods affect the performance. The effect of Regularization Coefficients is discussed later. \textcolor{black}{Then}, we analyze the influences of Graph Node Embedding Size. In Fig. \ref{fig:para}, we show the effect of parameters on average error distance, median error distance and max error distance \textcolor{black}{with training-70\% datasets in the first 3 networks (NYS, HK and Shanghai) as an example}. However, we mainly analyze the effect of parameters on average error distance because the optimized target is MSE loss. GNN-Geo did not specifically seek a minimum median error distance or max error distance. We pick the best performance for each parameter on the testing sets. And no matter how parameters vary, the average error distances in Fig. \ref{fig:para} are consistently smaller than baselines in previous works for all datasets. It verifies the effectiveness of GNN-Geo, empirically showing that explicitly modeling the network connectivity can greatly facilitate the geolocation task. {Finally, we also investigate how different GNN basic models would affect the performance of GNN-Geo.}
    \subsubsection{Effect of the Depth of Message Passing Layers}

    %    \item[3] When varying the number of propagation layers, GNN-Geo-v4 is consistently superior to baselines across three datasets. It again verifies the effectiveness of GNN-Geo-v4, empirically showing that explicit modeling of high-order connectivity can greatly facilitate the geolocation task.
    %    \end{itemize}
    The depth of the MP layers is actually the number of MP layers. From Fig. \ref{fig:para_layer_average}, the best layer numbers are 3, 2, 2 in NYS, HK and Shanghai, reflectively. The worst layer numbers are 1, 5 and 1 in NYS, HK and Shanghai, respectively. Compared to the worst layer numbers,  the improvements of the best layer numbers are 7.27\% \textasciitilde\ 16.55\%. Increasing the depth of GNN-Geo substantially enhances the geolocation performance. GNN-Geo-3 indicates GNN-Geo with three message propagation layers and similar notations for others. We can see that GNN-Geo-2, GNN-Geo-3 and GNN-Geo-4 achieve consistent improvement over GNN-Geo-1 across all datasets, which only consider the first-order neighbors only. When further stacking the propagation layer on the top of GNN-Geo-4, we find that GNN-Geo-5 leads to overfitting on the Hong Kong dataset. This might be caused by applying a too deep architecture that might introduce noises to the representation learning. The marginal improvements on the other two datasets verify that conducting 2-3 propagation layers are sufficient to model the network for IP geolocation.

    \subsubsection{Effect of Aggregation Methods}

    From Fig. \ref{fig:para_agg_average}, the best aggregation methods are Max, Mean, Mean in NYS, HK and Shanghai, respectively. The worst aggregation methods are Sum, Max, Max in NYS, HK and Shanghai, respectively. Compared to the worst aggregation methods,  the improvements of the best aggregation methods are 9.58\%, 1.92\%, 18.95\%, respectively. Neighborhood aggregation is a key operation, which distinguishes GNN from MLP as we discussed in section \ref{sec:compare}. Theoretically, all three methods have their limitation: 1) both Mean and Sum cannot treat each neighbor differently; 2) Max may only lose some neighbors' useful information. Previous work \cite{xu2018powerful} has pointed out that Sum may be more powerful than Mean and Max for graphs without attributes. However, in three datasets of this paper, Mean is the best or the second-best, while Max is a little worse than Mean. Max could be the worst or the best in different datasets. The attributed graph is more complex because the attributes of edges and nodes are also influencing the performance besides the network structure. The best aggregation methods in attributed graphs for geolocation still need more depth analysis.

    \subsubsection{Effect of Graph Node Embedding Size}

    From Fig. \ref{fig:para_nodeem_average}, all the best graph node embedding sizes are 128 in NYS, HK and Shanghai. All the worst graph node embedding sizes are 256 in NYS, HK and Shanghai. Compared to the worst graph node embedding sizes,  the improvements of the best graph node embedding sizes are 5.71\%, 5.68\%, 14.49\%, respectively. The influence patterns of Graph Node Embedding Size are stable for all three datasets. Increasing embedding size brings more representation power to the MP layers by introducing more complexity. This enables the model to learn more complicated patterns. So embedding size 128 performs better than 32 and 64 in all datasets. However, a too large embedding size like 256 could disturb the model with noises and result in overfitting, especially if there are not enough data instances for training.

\begin{figure*}[htbp]
        \centering
        \subfigure[Training Dataset/Full Dataset = 70\%]{
        \begin{minipage}[t]{0.48\linewidth}
        \centering
        \includegraphics[width=3in]{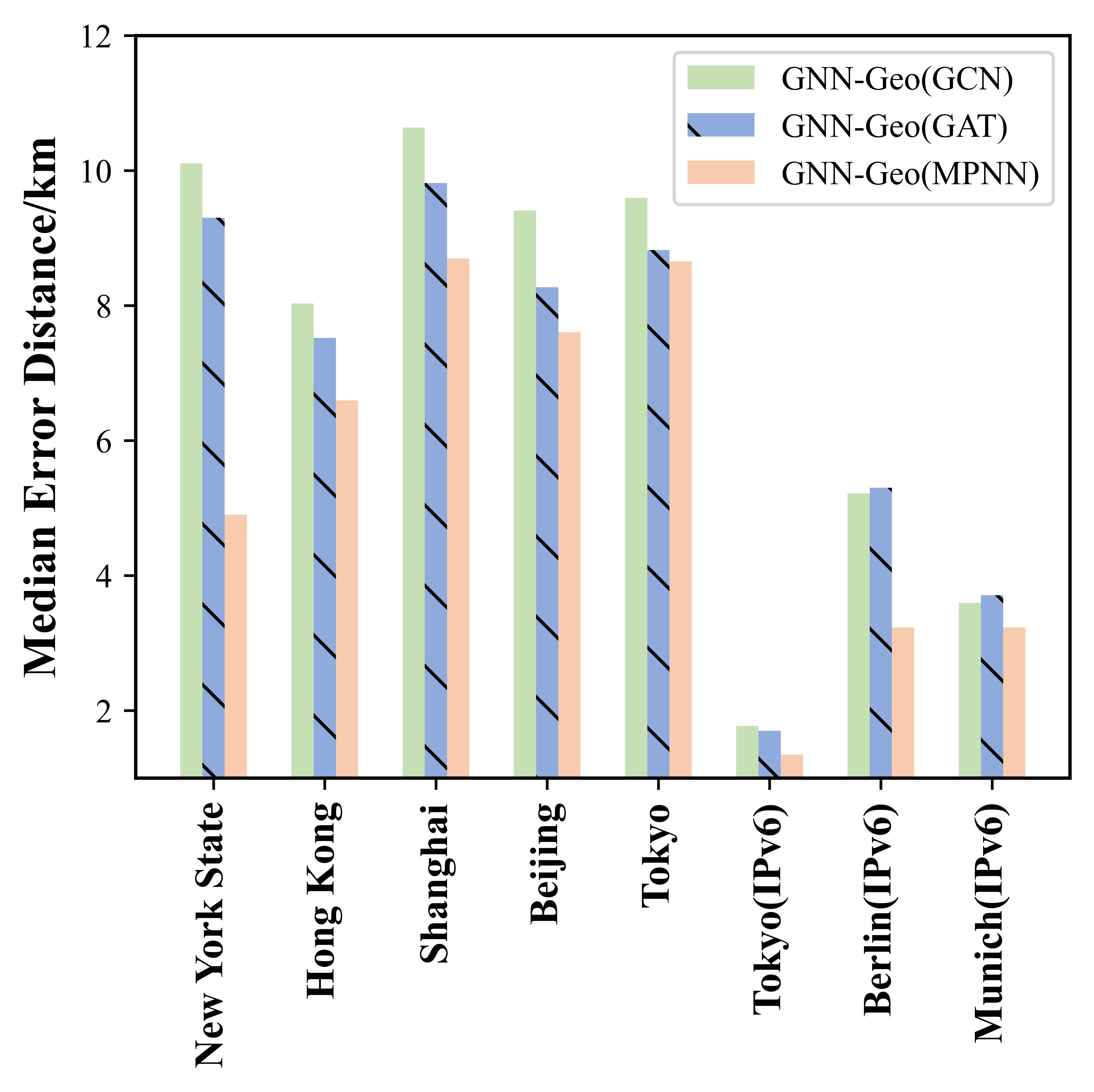}
        %\caption{fig1}
        \end{minipage}%
        \label{fig:gnns_train70}
        }%
        \subfigure[Training Dataset/Full Dataset = 10\%]{
        \begin{minipage}[t]{0.48\linewidth}
        \centering
        \includegraphics[width=3in]{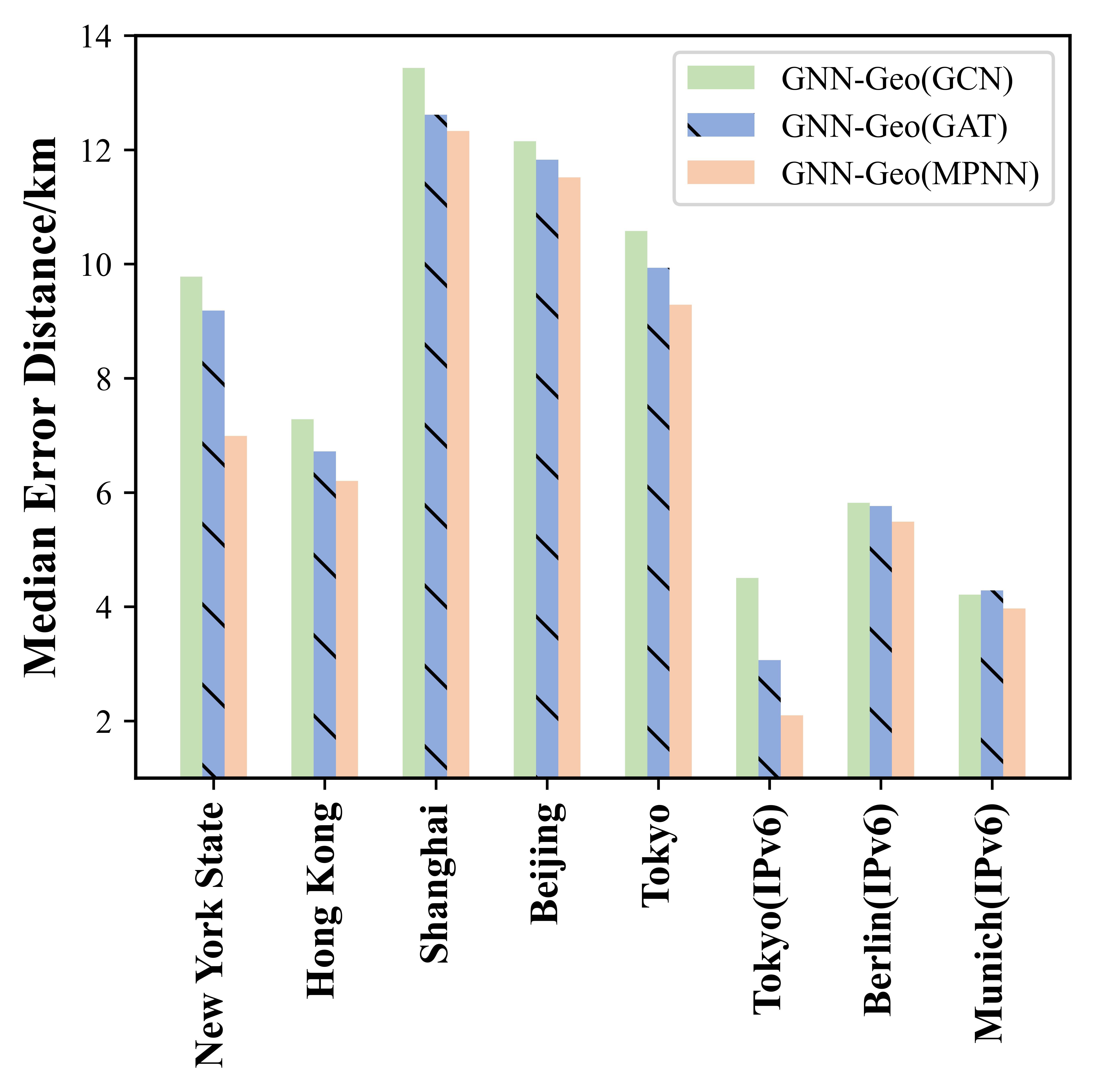}
        %\caption{fig2}
        \end{minipage}%
        \label{fig:gnns_train10}
        }
        \centering
        \caption{\textcolor{black}{Median Error Distances of Different GNN Basic Models in 8 Networks}}
        \label{fig:gnns}
\end{figure*}

    \subsubsection{Effect of Regularization Coefficients}

    From Fig. \ref{fig:para_reg_average}, the best Regularization Coefficients are 0.0005, 0.001, 0.001 in NYS, HK and Shanghai, respectively. The worst Regularization Coefficients are 0.01, 0.01, 0.0005 in NYS, HK and Shanghai, respectively. Compared to the worst Regularization Coefficients, the improvements of the best Regularization Coefficients are 31.90\%, 1.47\%, 6.36\%. $L_{2}$ Regularization is the main method in GNN-Geo to overcome overfitting. However, too larger Regularization Coefficients could lead to underfitting. We find that the influence of Regularization Coefficients varies significantly in different datasets. The Regularization Coefficient is very important to NYS dataset since the improvement is larger than all the other parameters discussed in this section. However, the Regularization coefficient is the least influential parameter in the Hong Kong dataset, compared with the other parameters. Generally, the Regularization Coefficients 0.001 should be tried first because it is the best and second-best in all datasets.

    \subsubsection{Effect of Regularization Coefficients}

    From Fig. \ref{fig:para_reg_average}, the best Regularization Coefficients are 0.0005, 0.001, 0.001 in NYS, HK and Shanghai, respectively. The worst Regularization Coefficients are 0.01, 0.01, 0.0005 in NYS, HK and Shanghai, respectively. Compared to the worst Regularization Coefficients, the improvements of the best Regularization Coefficients are 31.90\%, 1.47\%, 6.36\%. $L_{2}$ Regularization is the main method in GNN-Geo to overcome overfitting. However, too larger Regularization Coefficients could lead to underfitting. We find that the influence of Regularization Coefficients varies significantly in different datasets. The Regularization Coefficient is very important to NYS dataset since the improvement is larger than all the other parameters discussed in this section. However, the Regularization coefficient is the least influential parameter in the Hong Kong dataset, compared with the other parameters. Generally, the Regularization Coefficients 0.001 should be tried first because it is the best and second-best in all datasets.

    \subsubsection{\textcolor{black}{Effect of Different basic GNN models}}
    \textcolor{black}{In section 5 and all previous experiments of Section 6, we select MPNN as the basic GNN model for message passing layers. However, besides MPNN, there are still other famous basic GNN models such as GCN \cite{1st_GCN} and GAT \cite{velivckovic2017graph}. Can they be better than MPNN for IP geolocation tasks? To answer this question, GCN and GAT are also used for message passing layers, which are referred as GNN-Geo-GCN and GNN-Geo-GAT, respectively. We select training-70\% and training-10\% to test their performances. After parameters tuning, the average error distances and median error distances of GNN-Geo (GCN), GNN-Geo (GAT) and GNN-Geo (MPNN) are shown in Fig. \ref{fig:gnns}. From Fig. \ref{fig:gnns}, we can see for both average and median error distances, GNN-Geo (MPNN) outperforms the other two methods on training-70\% and training-10\% datasets. This validates the importance to explicitly model edge features for IP geolocation. GNN-Geo (GAT) is usually better than GNN-Geo (GCN), which may indicate that some neighbors should be more important than others, and attention mechanism should also be tried in improving GNN-Geo in the future. Besides, compared with train-10\%, the advantages of MPNN are usually clearer in training-70\% datasets. This may show that GNN-Geo (MPNN) is easier to overfit than other two methods on smaller training datasets.}

    \section{Discussion}
    \label{sec:discussion}
    In this paper, we mainly test whether GNN-Geo's accuracy outperforms previous fine-grained IP geolocation methods. From the experiments in Section \ref{sec:results_comparison}, we can see that GNN-Geo is more accurate than previous methods due to its stronger modeling ability on non-linear relationships hidden in computer network measurement data. In this section, we will discuss what improvements GNN-Geo still needs to make it more practical for industrial-scale IP geolocation tasks.

    \textbf{Increasing Accuracy with fewer Landmarks}. GNN is a deep learning-based method. Like other deep learning-based methods, its performance is much easier to be affected than rule-based methods when the landmarks or training data instances are too few \textcolor{black}{(as shown in Section 6.7)}. This is because deep learning methods are easier to overfit on sparse datasets. Thus, we need to consider how to improve GNN-Geo's performance in areas where the landmarks are hard to collect.

    \textbf{Injecting explainability into GNN-Geo} Compared with rule-based methods like SLG, deep learning-based methods often lack explainability for results. For example, for a target IP address in the Shanghai dataset, after geolocating it with SLG, it is easier to analyze why the error distance of this target is large. It is usually because its shortest delay does not come from its nearest landmark. However, it is hard to explain why GNN-Geo's error distance for this target is large. We need to discuss how to inject explainability into GNN-Geo's results for IP geolocation service users.

    %\textcolor[rgb]{0,0,0}{\textbf{Lower the Cost on GPU Computing Resoureces}. Compared with rule-based methods, deep learning-based methods often need GPU (graph processor unit) computing resources to acclerate the training processes. And MPNN-Geo need larger computing resources than MLP-Geo because it needs to learn more parameters to get the best results. This would limit MPNN-Geo's application for large-scale IP geolocation tasks. Thus we need to simplify the design and trainging process of MPNN-Geo with comparable accuracy.}

    %\textcolor[rgb]{0,0,0}{\textbf{Leverage Attention Mechanism for Feature Learning}. When forming the representation of each IP address, MPNN-Geo does not specially distinguish its neighbor's influence. In fact, it is possible that different neighours have different affect. For example, some neighhors could be much closer to the target IP addresses. We can use attention mechanism to differeniate the neighbors' influnece and give more weights to some neighbors to increase the preformance and explainability. GAT proposed how to leverage attention mechanism for node feature learning. We could consider how to add attention mechanism into MPNN-Geo for both node and edge feature learning. In fact, MPNN-Geo is a basic GNN model. In this paper, MPNN-Geo just show the great potential of GNN for fine-grained IP geolocation. Besides attention mechanism, researchers could improve it with various recent findings in GNN areas to gain better performance.}

    \section{Conclusion}
    \label{sec:conclusion}
    In this work, we first formulate the measurement-based fine-grained IP geolocation task into a semi-supervised attributed-graph node regression problem. Subsequently, we \textcolor{black}{discuss why GNN is worth to be tested for IP geolocation}. Then, we propose a GNN-based fine-grained IP geolocation framework (GNN-Geo) to solve the node regression problem. The framework consists of a preprocessor, an encoder, MP layers and a decoder. Moreover, we ease the convergence problem of GNN-based IP geolocation methods by scaling the range of output as well as combining the batch normalization and Sigmoid functions into the decoder. Finally, the experiments in \textcolor{black}{8 different real-world IPv4/IPv6} datasets verify the generalization capabilities of GNN-Geo for fine-grained IP geolocation.\\
    \indent
    The main advantages of GNN-Geo are the strong modeling ability on the computer network measurement data and the powerful extraction ability of non-linear relationships. Compared with rule-based methods, its main disadvantage is explainability, which is also a common problem for deep learning-based methods. Compared with the previous deep learning-based methods, the stronger modeling ability of GNN-Geo also needs more computing resources for training. In future work, we plan to increase GNN-Geo's accuracy with fewer landmarks, inject explainability into GNN-Geo, lower the training cost of GNN-Geo and improve GNN-Geo with various mechanisms like attention.\\

%\appendices
%% you can choose not to have a title for an appendix
%% if you want by leaving the argument blank
%\section{}

\ifCLASSOPTIONcompsoc
  \section*{Acknowledgments}
\else
  \section*{Acknowledgment}
\fi

This work was supported by \textcolor{black}{the National Key R\&D Program of China (Grant No. 2022YFB3102904),} the National Natural Science Foundation of China (Grant No.U1804263, 61872448 and 62172435), and Zhongyuan Science and Technology Innovation Leading Talent Project of China (Grant No. 214200510019).

%\ifCLASSOPTIONcaptionsoff
%  \newpage
%\fi

%\begin{thebibliography}{1}

%\end{thebibliography}
\footnotesize
\bibliographystyle{ieeetr}
\bibliography{ref2}
\vskip 22mm
\begin{IEEEbiography}[{\includegraphics[width=1in,height=1.25in,clip,keepaspectratio]{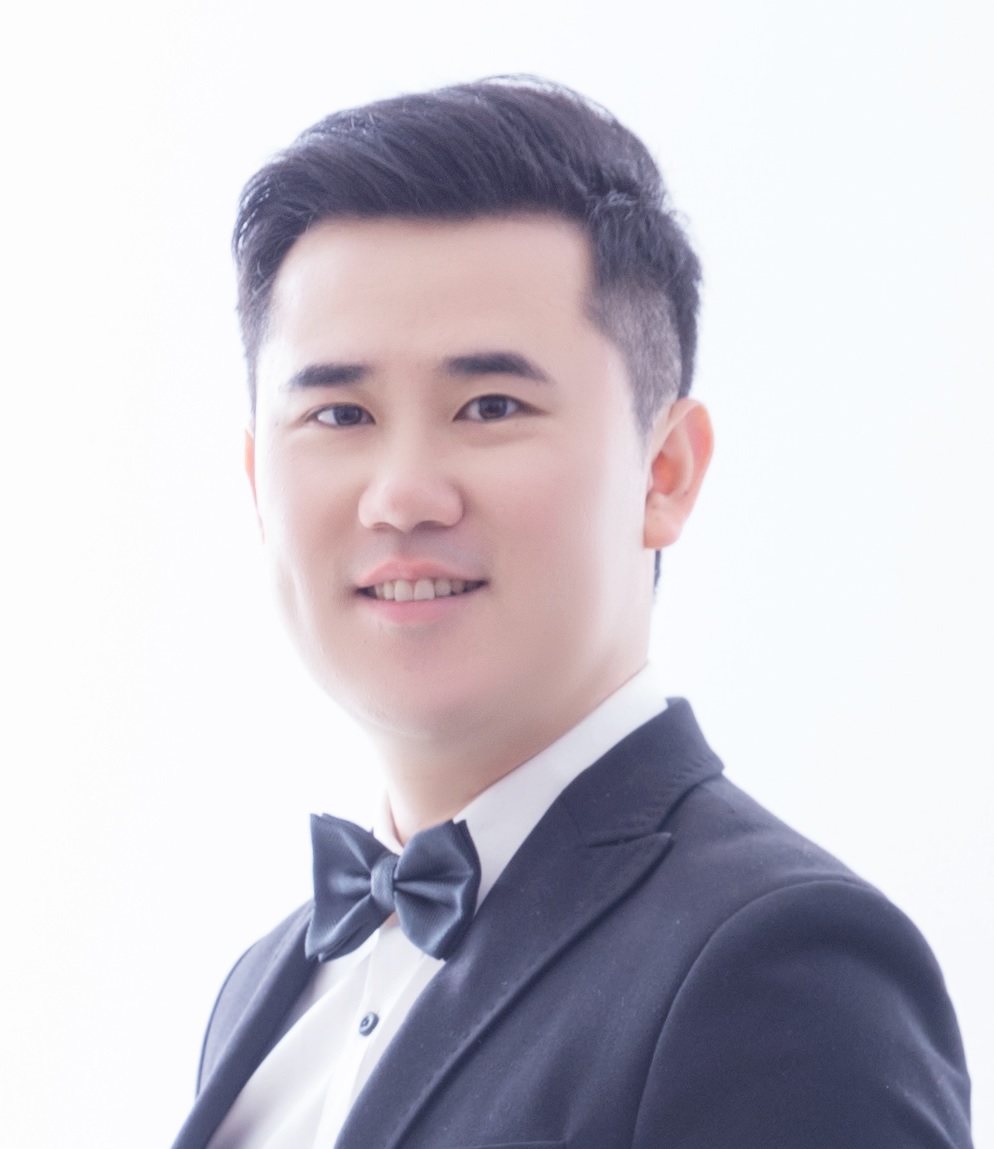}}]{Shichang Ding}
received the Ph.D. degree from University of G\"ottingen, Germany, in 2020. He is currently a lecturer at State Key Laboratory of Mathematical Engineering and Advanced Computing. His research interests include cyberspace surveying and mapping, graph deep learning, and social computing.
\end{IEEEbiography}
\vspace{-0.2cm}
\begin{IEEEbiography}[{\includegraphics[width=1in,height=1.25in,clip,keepaspectratio]{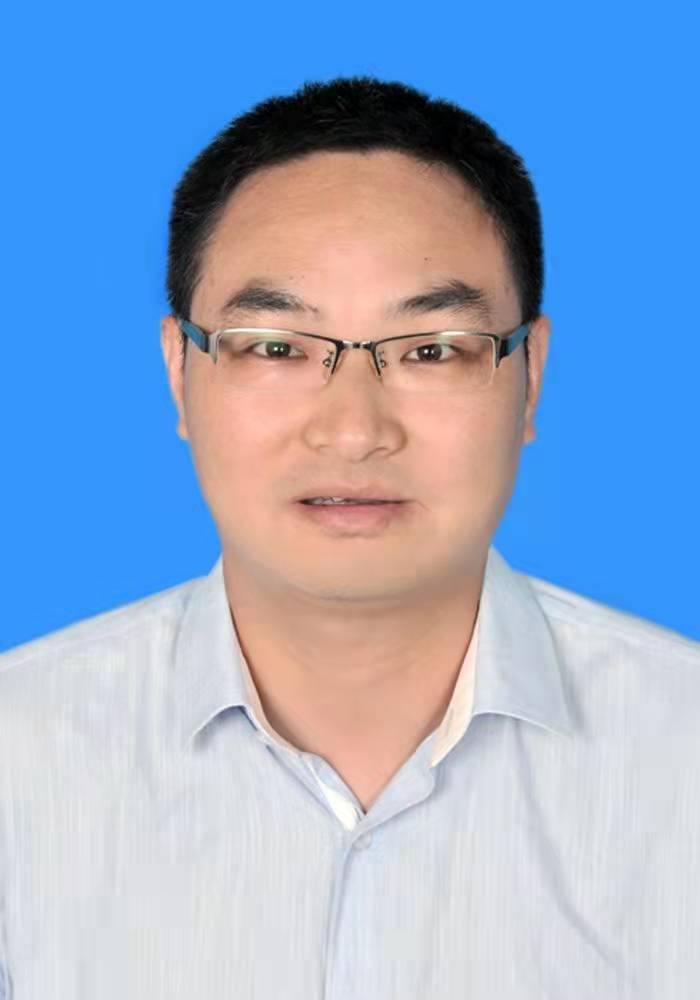}}]{Xiangyang Luo}
received the Ph.D. degree from Information Engineering University, China, in 2010. He is currently a Professor and a Ph.D. Supervisor with State Key Laboratory of Mathematical Engineering and Advanced Computing. He has authored or coauthored more than 150 refereed international journal and conference papers. His research interests include multimedia security and cyberspace surveying and mapping.
\end{IEEEbiography}
\vspace{-0.2cm}
\begin{IEEEbiography}[{\includegraphics[width=1in,height=1.25in,clip,keepaspectratio]{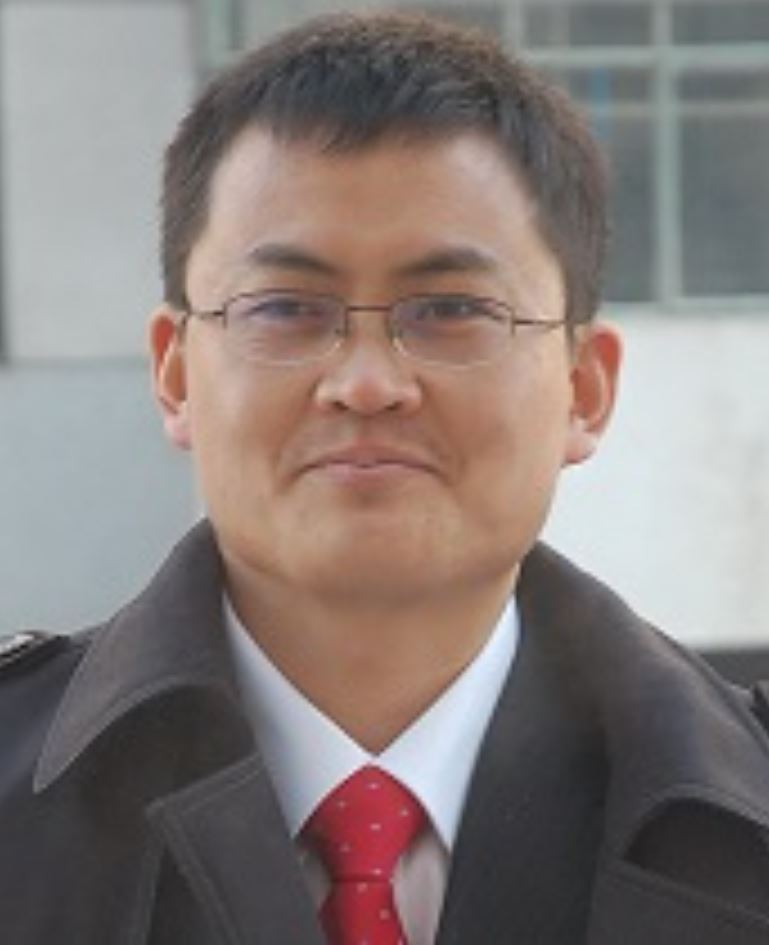}}]{Jinwei Wang} received the Ph.D. degree in information security from the Nanjing University of Science and Technology, China, in 2007. He is currently a Professor at the Nanjing University of Information Science and Technology. He has published over 50 papers. His research interests include multimedia copyright protection, multimedia forensics, multimedia encryption, and data authentication.
\end{IEEEbiography}
\vspace{-0.2cm}
\begin{IEEEbiography}[{\includegraphics[width=1in,height=1.25in,clip,keepaspectratio]{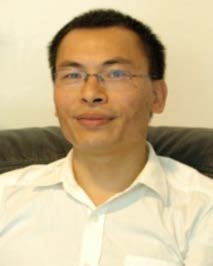}}]{Xiaoming Fu} received the PhD degree in computer science from Tsinghua University, Beijing, China in 2000. He was then a research staff at Technical University of Berlin until joining the University of G\"ottingen, Germany in 2002, where he has been a professor in computer science and heading the computer networks group since 2007. He has spent research visits at Cambridge, Columbia, UCLA, Tsinghua University, Uppsala, and UPMC, and is an IEEE fellow and distinguished lecturer. His research interests include Internet-based systems, applications, and social networks. He is currently an editorial board member of IEEE Communications Magazine, IEEE Transactions on Network and Service Management, Elsevier Computer Networks, and Computer Communications, and has published over 150 peer-reviewed papers in renowned journals and international conference proceedings.
\end{IEEEbiography}

\clearpage
\end{document}